\documentclass[letterpaper]{article}
\usepackage[preprint]{aaai2027}
\usepackage[hyphens]{url}
\usepackage{graphicx}
\graphicspath{{fig2/}}
\usepackage{array}
\usepackage{adjustbox}
\usepackage{amsmath}
\usepackage{booktabs}
\usepackage{algorithm}
\usepackage{algorithmic}
\usepackage{multirow}
\usepackage{subcaption}
\usepackage{placeins}
\usepackage{xcolor}
\usepackage{amssymb}
\usepackage{natbib}
\usepackage{pifont}

\newcommand{\SASMatchedCueControl}{60.7}
\newcommand{\SASMatchedCueFull}{68.9}
\newcommand{\SASMatchedCueGain}{+8.2}
\newcommand{\SASMatchedCueGainCI}{[-3.2,+19.5]}
\newcommand{\SASMatchedPersistenceGain}{-0.05}

\title{EmoWorld: A Decoupled Affective Field for Controllable Emotional Video Generation}

\author {
   Bingyuan Wang\textsuperscript{\rm 1}, Baistan Zhyldyzbekov\textsuperscript{\rm 1}, Kunyu Feng\textsuperscript{\rm 1}, Zeyu Wang\textsuperscript{\rm 1,\rm 2}\corresponding
}
\affiliations {
   \textsuperscript{\rm 1}The Hong Kong University of Science and Technology (Guangzhou) \\
   \textsuperscript{\rm 2}The Hong Kong University of Science and Technology \\
   zeyuwang@ust.hk
}

\begin{document}

\maketitle

\begin{abstract}
Emotion shapes how viewers interpret a scene, yet existing video
generators entangle global atmosphere, affect-bearing semantic cues, and temporal progression within a single text condition. We present EmoWorld, a framework that decouples these factors within a frozen flow-matching video diffusion transformer (Video DiT). A one-time preparation stage extracts layer-specific affect directions and a reusable cue library from geometry-preserving neutral and emotion-edited panoramas. At inference, Visual Atmosphere Steering (VAS) injects atmosphere directions into hidden states, Semantic Affective Steering (SAS) isolates a separately scalable prompt residual for semantic cues, and Temporal Affective Steering (TAS) interpolates endpoint residual fields across denoising and video time. On Wan2.2, VAS improves target-emotion alignment by 19\% while reducing a temporal-fluctuation proxy by 48\%; SAS improves target-emotion alignment by 37\% and increases detected affect-bearing cues by 36\%; and TAS improves transition monotonicity by 15\% over the strongest baseline. EmoWorld is evaluated across 27 emotion categories in text-to-video and image-to-video settings, demonstrates portability across multiple Video-DiT backbones, and supports camera-conditioned composition without updating generator parameters.
\end{abstract}

\begin{figure*}[t]
  \centering
  \includegraphics[width=\textwidth]{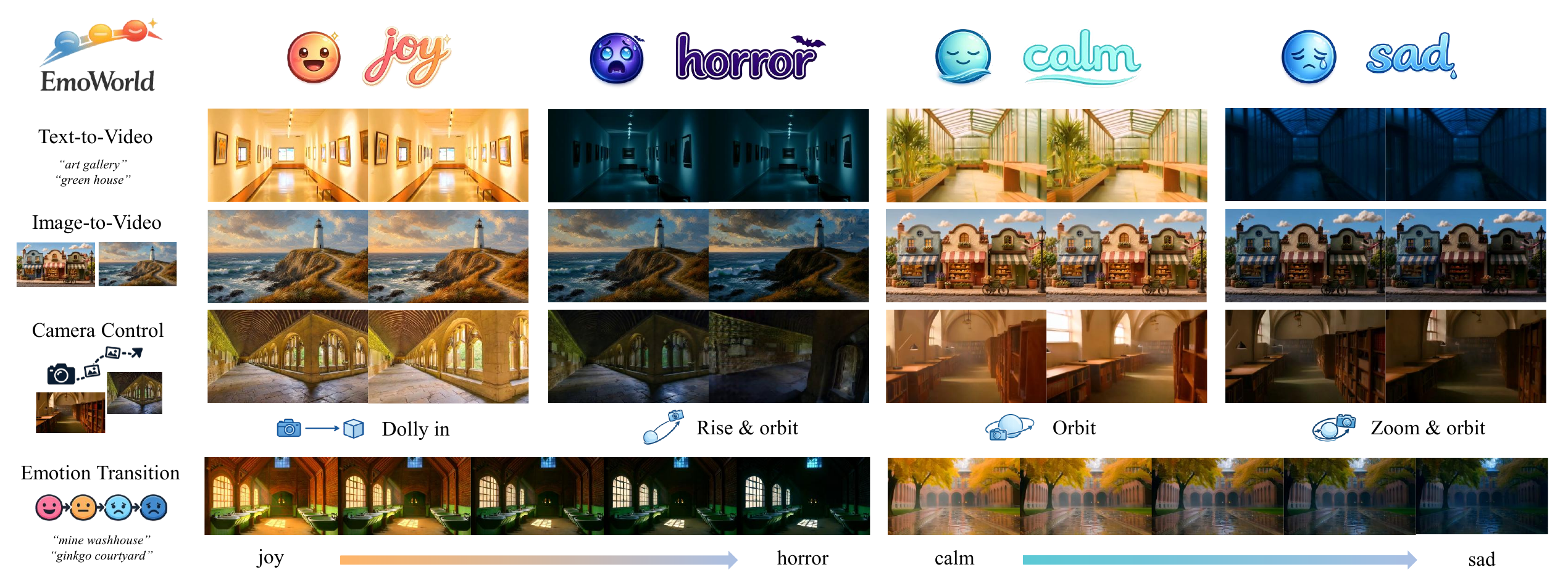}
  \vspace{-1em}
  \caption{\textbf{EmoWorld capabilities.} Text-to-video and
  image-to-video atmosphere control, semantic cue control,
  camera-conditioned generation, and temporal emotion transitions.}
  \label{fig:teaser}
\end{figure*}

\section{Introduction}

Emotion determines how viewers read a scene: the same street can
feel safe, nostalgic, melancholic, or ominous while its buildings
remain recognizable. This makes affective control valuable for
cinematic previsualization, story-driven generation, virtual
production, and interactive world authoring, where creators want
to direct how a place feels without replacing what the place is.
Affect, however, is not a single visual attribute. Global
illumination and color establish atmosphere; weather, vegetation,
decorations, and other localized cues carry semantic affect; and
changes over time determine emotional progression. Moreover, practical applications require diverse control modes, from text and image inputs to camera-conditioned composition and temporal emotion transitions, as previewed in Figure~\ref{fig:teaser}. Such complexities pose challenges for effective control: forcing these factors through a single emotion phrase and conditioning channel often produces generic color grading, weak semantic cues, or abrupt temporal changes. 

Modern text-to-video (T2V) and image-to-video (I2V) models
synthesize coherent scenes and motion~\cite{wan2025,cogvideox,
opensora,videopoet}, while controllable-video methods provide
handles for structure, appearance, editing, and camera
motion~\cite{ccedit2024,rave2024,motionctrl,camtrol2024}. However, these methods do not explicitly separate global atmosphere,
affect-bearing semantic cues, and temporal evolution. Meanwhile, video generation commonly leverages abundant image data through
joint training or image-model initialization~\cite{
ho2022video,blattmann2023stable}. Yet transferring paired affective images into scene-level affect control remains underexplored, while matched videos depicting the same scene under different emotions remain scarce.

We address these challenges with the \emph{EmoWorld} framework, which represents emotion control as a \emph{decoupled affective field} in a frozen Video DiT. We construct neutral and emotion-edited panorama pairs with preserved geometry as a
practical data source, and design a paired multimodal preparation stage to derive complementary feature-space directions from frozen Video-DiT probes and language-space cues from a vision-language model. At inference, EmoWorld retrieves scene-compatible cues to compose an emotion-augmented prompt and selects the corresponding precomputed affect directions for feature-space steering. 

We propose a set of three affective steering operators that share one frozen generator but act on complementary supports. Among them, VAS injects layer-specific affect directions into selected hidden states to establish scene-wide atmosphere; SAS decomposes prompt-induced prediction differences into a sparse, independently scalable residual for localized affective cues; and TAS treats endpoint emotions as boundary residual fields and interpolates their directions and magnitudes across denoising steps and video frames to generate smooth affect transitions. Additionally, a camera-conditioned variant combines prescribed viewpoint motion with scheduled atmosphere steering.

We evaluate EmoWorld's capability on atmosphere control across the complete 27-emotion taxonomy in both T2V and I2V settings, assess semantic cue control and temporal emotion transitions under controlled protocols, and demonstrate VAS portability across multiple Video-DiT backbones and camera-conditioned composition. On Wan2.2 T2V, VAS raises CLIP-Emo from 0.168 to
0.200 while reducing the temporal-fluctuation proxy from
1.22 to 0.63. In the cue-matched SAS evaluation, adding
SAS to Cue+VAS raises CLIP-Emo from 0.155 to 0.212 and increases the mean number of detected affect-bearing cues from 2.57 to 3.50 per sampled frame. TAS achieves a transition monotonicity of 0.788, compared with 0.687 for the strongest baseline.

Our contributions are threefold:
\begin{itemize}
    \item We propose the EmoWorld framework and formulate \emph{decoupled emotional video control},
    representing visual atmosphere, affect-bearing semantic cues,
    and temporal evolution as complementary components of a
    task-indexed affective field.

    \item We introduce a paired multimodal preparation procedure
    and two complementary operators for single-emotion control:
    VAS applies emotion- and layer-specific hidden-state steering,
    while SAS isolates a projected, sparse, and independently
    scalable prompt-residual correction.

    \item We introduce TAS, which generates emotion transitions
    through boundary-conditioned great-circle interpolation of
    endpoint residual directions, and provide a comprehensive
    evaluation covering different emotions, tasks and backbones, demonstrating superior results.
\end{itemize}

\section{Related Work}
\label{sec:related}

\paragraph{Controllable scene and video generation.}
Panoramic scene generation has progressed from single-panorama reconstruction to diffusion-based world creation and scene extension~\cite{automatic3dindoor2018,text2room,luciddreamer,wonderworld,holodreamer2026tvcg,dreamscene360,layerpano3d,panodreamer,matrix3d}. Recent surveys map the broader landscapes of diffusion-based visual art and controllable video generation~\cite{wang2025diffusion,ma2025controllable}. Modern video generators provide strong T2V and I2V backbones~\cite{wan2025,cogvideox,opensora,videopoet,hunyuanvideo}, while controllable methods introduce structural, camera, trajectory, and editing constraints~\cite{controlnet,controlvideo,motionctrl,camtrol2024,rave2024,ccedit2024}. Related systems support targeted content manipulation through DiT-based image editing, video inpainting for 4D creation, and efficient motion transfer~\cite{feng2025dit4edit,ma2025followcreation,ma2025followyourmotion}, while application-oriented work explores controllable immersive storytelling~\cite{wang2025magicscroll}. These works mainly control content, structure, or viewpoint; EmoWorld instead controls the affective interpretation of a scene.

\paragraph{Representation steering and training-free diffusion control.}
Activation steering controls high-level concepts through internal directions without retraining~\cite{repeng,turner2023activation,rimsky2024steering,linearrep2024,geometrytruth2024}. Related diffusion methods manipulate concept directions, selected features, attention, or flow-space residuals~\cite{conceptspaces2026,reins2026,lalqr2026,moft2024,stableflow2025,sega2023,freelsliders2025,flowedit2025,flowdirector2026,dynaedit2026}. EmoWorld extends this paradigm to video affect: our VAS steers internal atmosphere representations, SAS isolates semantic prompt residuals, and TAS interpolates endpoint residual fields over denoising and video time, rather than relying on frame-wise prompting~\cite{prompt2prog2025,tunerdit2026}.

\paragraph{Affective image and video generation.}
Affective image generation conditions diffusion models on emotion
categories, valence--arousal coordinates, editing prompts, or learned
adapters~\cite{emoedit2025,emoticrafter2025,make_me_happier2024,
emoagent2025,coemogen2025,uniemo2025,
epig2026,cogblender2026}. EmoSpace further studies fine-grained
affect control for panoramic scenes~\cite{emospace2026}. Compared
with the image literature, affective video generation remains less
developed and has largely focused on human-centric expression, such
as portrait animation~\cite{ma2025controllable,liu2025moee}. Recent work such as EmoVid extends emotion-conditioned generation
to broader video content~\cite{emovid2026}, whereas EmoWorld
decouples global atmosphere, localized semantic cues, and temporal
affect evolution within a frozen video generator.

\section{Method}

\begin{figure*}[t]
  \centering
  \includegraphics[width=\textwidth]{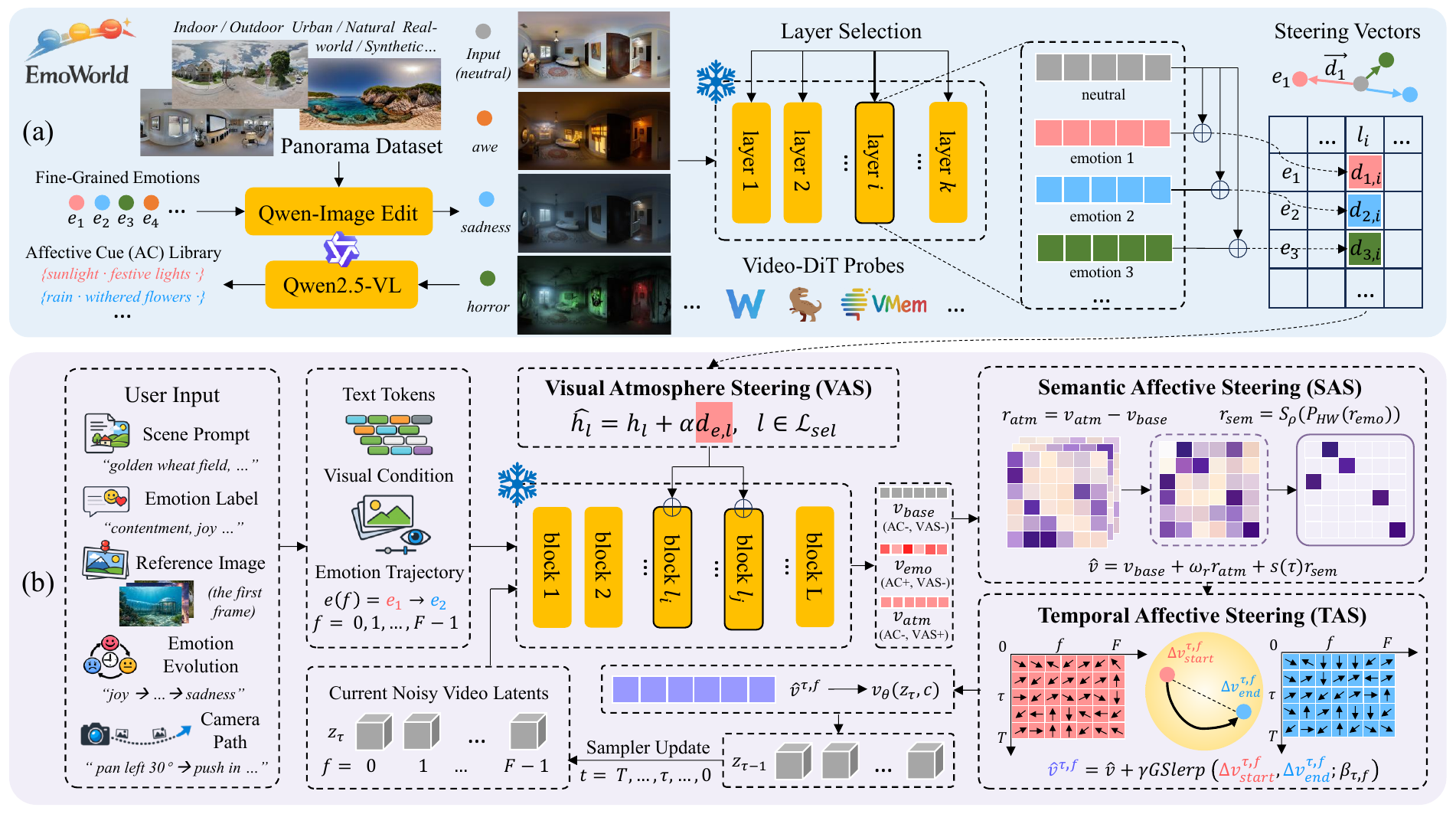}
  \caption{\textbf{EmoWorld overview.} (a) Geometry-preserving neutral and edited panorama pairs yield feature-space VAS vectors and a language-space affective cue library. (b) During inference, VAS, SAS, and TAS act on hidden states, prediction residuals, and frame-wise velocity fields, respectively, within a shared frozen Video DiT.}
  \label{fig:pipeline}
\end{figure*}

Figure~\ref{fig:pipeline} summarizes EmoWorld. A one-time paired preparation stage uses geometry-preserving emotional edits in two complementary ways: the frozen Video-DiT probes produce emotion- and layer-specific VAS steering vectors, while a frozen difference-aware vision--language model constructs an affective cue (AC) library. At inference, the framework retrieves scene-compatible cues from this library and composes the emotion-augmented text condition. VAS, SAS, and TAS act at three distinct points in the Video DiT computation: selected hidden states, residuals between complete velocity predictions, and frame-wise velocity fields before the sampler update. Each task activates only the required operators, as illustrated in the lower panel of Figure~\ref{fig:pipeline}.

\subsection{Video-DiT Inference and Operator Placement}
\label{sec:dit_inference}

At denoising step $q$, the noisy latent $\mathbf{x}_q\in\mathbb{R}^{C\times F\times H\times W}$ already contains all $F$ latent video frames. A complete conditional forward pass propagates the corresponding spatiotemporal tokens sequentially through the DiT blocks and an output head,
\begin{equation}
    \mathbf v^{q}(p;\mathcal A)
    =
    \Phi_{\theta}(\mathbf{x}_q,t_q,p;\mathcal A)
    \in\mathbb{R}^{C\times F\times H\times W},
    \label{eq:dit_forward}
\end{equation}
where $p$ denotes the text condition and $\mathcal A$ denotes an optional internal intervention. The block index is network depth, the frame index $f$ identifies slices within the same predicted velocity tensor, and the denoising index $q$ belongs to the outer sampler loop. After any task-specific velocity assembly, the sampler integrates the prediction to obtain
\begin{equation}
    \mathbf{x}_{q-1}
    =
    \operatorname{Update}(\mathbf{x}_q,\widehat{\mathbf v}^{q}).
    \label{eq:sampler_update}
\end{equation}
The same frozen weights are reused for all required branches. Static and semantic control use the base, cue-augmented, VAS-on base, and unconditional predictions $\{\mathbf v_b,\mathbf v_e,\mathbf v_b^{\mathrm{VAS}},\mathbf v_{\varnothing}\}$, which are assembled outside the backbone according to the active task. VAS acts inside selected blocks, SAS decomposes $\mathbf v_e-\mathbf v_b$ after complete forward passes, and TAS supplies a frame-wise transported residual before Eq.~\ref{eq:sampler_update}. Thus, the operators share one generator without forming a mandatory serial chain.

\subsection{Paired Affective Preparation}
\label{sec:paired_preparation}

For each extraction scene $i$ and emotion $e$, we use a neutral panorama $\mathbf x_i^{\mathrm{neu}}$ and a geometry-preserving emotional edit $\mathbf x_{i,e}^{\mathrm{emo}}$. The same pair supports two complementary representations of affect: a feature-space steering vector for VAS and a language-space cue description for SAS.

\paragraph{Feature-space steering vectors.}
Let $\mathbf v_\theta$ denote a pretrained flow-matching video DiT. We extract matched feature summaries at selected layers and compute
\begin{equation}
    \boldsymbol{\delta}_{e,l}^{(i)}
    =
    \mathbf f_{e,l}^{(i,\mathrm{emo})}
    -
    \mathbf f_{l}^{(i,\mathrm{neu})}.
    \label{eq:emotion_delta}
\end{equation}
For $N_e$ paired observations of emotion $e$, the emotion- and layer-specific steering vector is
\begin{equation}
    \mathbf d_{e,l}
    =
    \frac{
    \frac{1}{N_e}\sum_{i=1}^{N_e}\boldsymbol{\delta}_{e,l}^{(i)}
    }{
    \left\|\frac{1}{N_e}\sum_{i=1}^{N_e}\boldsymbol{\delta}_{e,l}^{(i)}\right\|_2+\epsilon
    }.
    \label{eq:steering_vector}
\end{equation}
The resulting collection contains one normalized steering vector for each emotion and selected layer. Construction requires an offline editing and feature-probing pass, but optimizes no video-generator parameters. Static panoramas are suitable because VAS targets layer-wise changes in visual atmosphere rather than temporal motion.

\paragraph{Language-space affective cues.}
We additionally query a frozen difference-aware vision--language model (Qwen2.5-VL in our implementation) with the neutral and edited image pair and the emotion label. The instruction asks for visual changes that communicate the target affect while excluding preserved scene identity, geometry, and viewpoint. The response is normalized into atomic descriptors of atmosphere and scene-compatible semantic cues, such as warm illumination, festive lights, rain, withered vegetation, or ominous shadows. After deduplication and grouping by emotion, these descriptors form an affective cue library
\begin{equation}
    \mathcal C_e=\{c_{e,1},\ldots,c_{e,M_e}\}.
    \label{eq:cue_library}
\end{equation}
Given a base scene prompt $p_b$ and target emotion $e$, we retrieve a compatible subset and compose the emotion-augmented prompt
\begin{equation}
\begin{aligned}
\mathcal C_e(p_b)
&=
\operatorname{Retrieve}
\!\left(p_b,\mathcal C_e\right),
\\
p_e
&=
\operatorname{Compose}
\!\left(
    p_b,e,\mathcal C_e(p_b)
\right).
\end{aligned}
\label{eq:prompt_composition}
\end{equation}
The cue library and composed prompt set are frozen before matched evaluation. At video-generation time, the user supplies only the scene prompt, emotion label or trajectory, and optional reference-image or camera conditions; the edited panoramas are not inputs to the video generator. SAS does not use the VAS vectors directly, while TAS uses VAS-conditioned endpoint predictions in velocity space, so paired emotional videos are unnecessary.

\subsection{Visual Atmosphere Steering}
\label{sec:vas}

VAS instantiates the layer-wise component of the decoupled affective field. For emotion $e$ and selected DiT block $l$, let $\mathbf d_{e,l}$ denote the normalized steering vector from Eq.~\ref{eq:steering_vector}. At a registered feature hook $r$, VAS broadcasts this vector over the corresponding video tokens:
\begin{equation}
    \widehat{\mathbf{h}}_{l,r}^{\,q,f}
    =
    \mathbf{h}_{l,r}^{\,q,f}
    +
    \gamma_{e,l,r}^{\,q,f}\,\mathbf{d}_{e,l},
    \label{eq:vas_injection}
\end{equation}
where $q$ indexes the denoising step, $f$ indexes the latent frame, and $\gamma_{e,l,r}^{\,q,f}$ is a route-specific gain. At self-attention inputs, the gain combines user-specified strength with depth, denoising-step, and frame schedules; late cross-attention outputs use separate layer- and emotion-dependent coefficients.

Prompt-only conditioning must encode scene content and affect through the same text pathway. VAS instead modifies internal representations along an affective steering vector already present in the model, primarily changing global lighting, color temperature, contrast, saturation, and related atmosphere cues. Localized semantic cues are handled separately by SAS.

Let $\mathbf{v}_{b}$ be the prediction under the base scene prompt with VAS disabled and let $\mathbf{v}_{b}^{\mathrm{VAS}}$ use the same prompt with VAS enabled. The corresponding output-space residual is
\begin{equation}
    \mathbf{r}_{\mathrm{atm}}
    =
    \mathbf{v}_{b}^{\mathrm{VAS}}-\mathbf{v}_{b}.
    \label{eq:vas_residual}
\end{equation}
With unconditional prediction $\mathbf{v}_{\varnothing}$ and classifier-free guidance (CFG) scale $w$, the VAS-only configuration is $\mathbf{v}_{\varnothing}+w(\mathbf{v}_{b}^{\mathrm{VAS}}-\mathbf{v}_{\varnothing})$. Prompt+VAS denotes the matched condition that applies the same library-composed emotion-augmented prompt as Prompt-only while enabling VAS.

\subsection{Semantic Affective Steering}
\label{sec:sas}

SAS controls affect-bearing semantics through a spatially decomposed prompt-velocity residual. The emotion-augmented prompt $p_e$ combines the base scene prompt $p_b$, the target emotion, and scene-compatible descriptors retrieved from the offline cue library in Eq.~\ref{eq:prompt_composition}. These descriptors can introduce rain, withered vegetation, warm decorations, ominous shadows, and other localized affective cues. We decompose the resulting prompt-induced velocity difference into a per-frame spatially constant component and a zero-mean spatial component. The projector $\mathcal P_{HW}$ removes the former, and the magnitude-based operator $\mathcal S_{\rho}$ retains the strongest coordinates of the latter. This yields a sparse semantic correction that is independently scalable from VAS atmosphere steering; it is neither an object mask nor a segmentation map.

For the same latent, denoising step, and non-text conditioning, shared-weight forward passes produce $\mathbf v_b$, $\mathbf v_e$, $\mathbf v_{\varnothing}$, and $\mathbf v_b^{\mathrm{VAS}}$ from the base prompt, cue-augmented prompt, unconditional text condition, and VAS-enabled base prompt, respectively. The atmosphere branch remains independently controlled through $\mathbf r_{\mathrm{atm}}=\mathbf v_b^{\mathrm{VAS}}-\mathbf v_b$. Every symbol therefore denotes a complete DiT prediction rather than the output of an individual transformer block.

\paragraph{Per-frame spatial projection.}
For a prompt residual $\mathbf{r}=\mathbf{v}_e-\mathbf{v}_b\in\mathbb{R}^{C\times F\times H\times W}$, we remove the spatial mean independently for every channel and latent frame:
\begin{equation}
    [\mathcal{P}_{HW}(\mathbf{r})]_{c,f,h,w}
    =
    r_{c,f,h,w}
    -
    \frac{1}{HW}\sum_{h'=1}^{H}\sum_{w'=1}^{W}r_{c,f,h',w'}.
    \label{eq:sas_project}
\end{equation}
The dense projected residual has zero spatial mean for every $(c,f)$, removing frame-wise global bias before nonlinear sparsification.

\paragraph{Coordinate sparsification and schedule.}
We retain the largest-magnitude fraction $\rho$, where $0<\rho\leq1$:
\begin{equation}
    \mathbf{r}_{\mathrm{sem}}
    =
    \mathcal{S}_{\rho}\!\left(\mathcal{P}_{HW}(\mathbf{v}_e-\mathbf{v}_b)\right),
    \label{eq:sas_sparse}
\end{equation}
where $\mathcal{S}_{\rho}$ applies a single magnitude threshold over the full $C\times F\times H\times W$ tensor and retains the top-$\rho$ fraction of residual coordinates.

We modulate this correction with a warmup--hold--fade schedule $s(\xi_i)$ over normalized denoising progress $\xi_i=i/N$. The branch is gradually activated, held through the middle denoising interval, and faded near the end; the exact piecewise definition and boundary cases are provided in Appendix~\ref{sec:sas_tensor_cost}.

\paragraph{Four-pass assembly.}
The final velocity is
\begin{equation}
    \widehat{\mathbf{v}}_i
    =
    \mathbf{v}_{\varnothing}
    + w(\mathbf{v}_b-\mathbf{v}_{\varnothing})
    + \lambda_{\mathrm{sem}}s(\xi_i)\mathbf{r}_{\mathrm{sem}}
    + w\mathbf{r}_{\mathrm{atm}},
    \label{eq:sas_combine}
\end{equation}
where $w$ is the CFG scale and $\lambda_{\mathrm{sem}}$ independently controls the semantic branch. Setting $\lambda_{\mathrm{sem}}=0$ recovers base-prompt VAS generation. With VAS disabled, $\mathcal P_{HW}$ and $\mathcal S_{\rho}$ replaced by the identity, $s(\xi_i)=1$, and $\lambda_{\mathrm{sem}}=w$, Eq.~\ref{eq:sas_combine} reduces to standard cue-prompt CFG. SAS therefore adds residual isolation, scheduling, and independent scaling without changing the generator.

\subsection{Temporal Affective Steering}
\label{sec:tas}

TAS instantiates the temporal component through \emph{boundary-conditioned interpolation in the generative velocity field}. Flow matching evolves the latent over denoising steps $q$, while each latent $\mathbf x_q$ contains all video frames $f$; conditional velocities therefore live on the product domain $(q,f)$. Start- and end-emotion passes with their corresponding VAS vectors, together with a shared negative-conditioning pass, define endpoint residual fields
\begin{equation}
    \Delta\mathbf{v}_{e_i}^{\,q,f}
    =
    \mathbf{v}_{e_i}^{\,q,f}
    -
    \mathbf{v}_{\varnothing}^{\,q,f},
    \qquad i\in\{1,2\}.
    \label{eq:velocity_delta}
\end{equation}
Each residual describes how an endpoint emotion redirects frame $f$ at denoising step $q$. A narrative profile $\bar{\beta}_f$ specifies the desired video-time progression; relaxation over denoising steps and smoothing across neighboring frames produce the transition-coordinate field
\begin{equation}
    \beta:(q,f)\longmapsto\beta_{q,f}\in[0,1].
    \label{eq:tas_coordinate_field}
\end{equation}
Thus, every pair $(q,f)$ receives a continuous coordinate between the endpoint fields, explicitly separating the outer sampler trajectory from the internal video timeline.

For one complete latent frame, let $\mathbf z_i=\operatorname{vec}(\Delta\mathbf v_{e_i}^{\,q,f})$, $n_i=\|\mathbf z_i\|_2$, and $\mathbf u_i=\mathbf z_i/n_i$. Linear interpolation follows the chord between $\mathbf z_1$ and $\mathbf z_2$ and can shrink or cancel the residual when their directions differ. TAS instead follows the shortest great-circle path between normalized residual directions on the unit hypersphere and interpolates their magnitudes separately. With $\theta=\arccos(\operatorname{clamp}_{[-1,1]}\langle\mathbf u_1,\mathbf u_2\rangle)$, the nondegenerate case is
\begin{equation}
\begin{split}
    \operatorname{GSlerp}_{\beta}(\mathbf z_1,\mathbf z_2)
    ={}&
    \left[
    \frac{\sin((1-\beta)\theta)}{\sin\theta}\mathbf u_1
    +
    \frac{\sin(\beta\theta)}{\sin\theta}\mathbf u_2
    \right] \\
    &\cdot\left[(1-\beta)n_1+\beta n_2\right].
    \label{eq:tas_slerp}
\end{split}
\end{equation}
Near-zero, parallel, and antipodal cases are handled with
numerical fallbacks in implementation. The transported frame-wise residual field is
\begin{equation}
    \Delta\mathbf v_{\mathrm{TAS}}^{\,q,f}
    =
    \operatorname{unvec}\!\left[
      \operatorname{GSlerp}_{\beta_{q,f}}(\mathbf z_1,\mathbf z_2)
    \right],
    \label{eq:tas_transport}
\end{equation}
and the velocity passed to the sampler is
\begin{equation}
    \mathbf{v}_{\mathrm{TAS}}^{\,q,f}
    =
    \mathbf{v}_{\varnothing}^{\,q,f}
    + w\eta_{\mathrm{TAS}}\,
      \Delta\mathbf v_{\mathrm{TAS}}^{\,q,f},
    \label{eq:tas}
\end{equation}
where $w$ is the CFG scale and $\eta_{\mathrm{TAS}}$ controls residual amplification. The sampler then integrates the interpolated velocity field through Eq.~\ref{eq:sampler_update}; TAS therefore creates affect progression during generation rather than blending prompts or rendered frames after generation.

\subsection{Camera-Aware Affective Control}
\label{sec:camera}

A prescribed camera trajectory can control viewpoint motion while a temporal schedule selects VAS steering vectors. Their composition enables affect-aware shot planning while preserving separate control over viewpoint and atmosphere.

\section{Experiments}

\begin{figure*}[t]
\centering
\includegraphics[width=\textwidth]{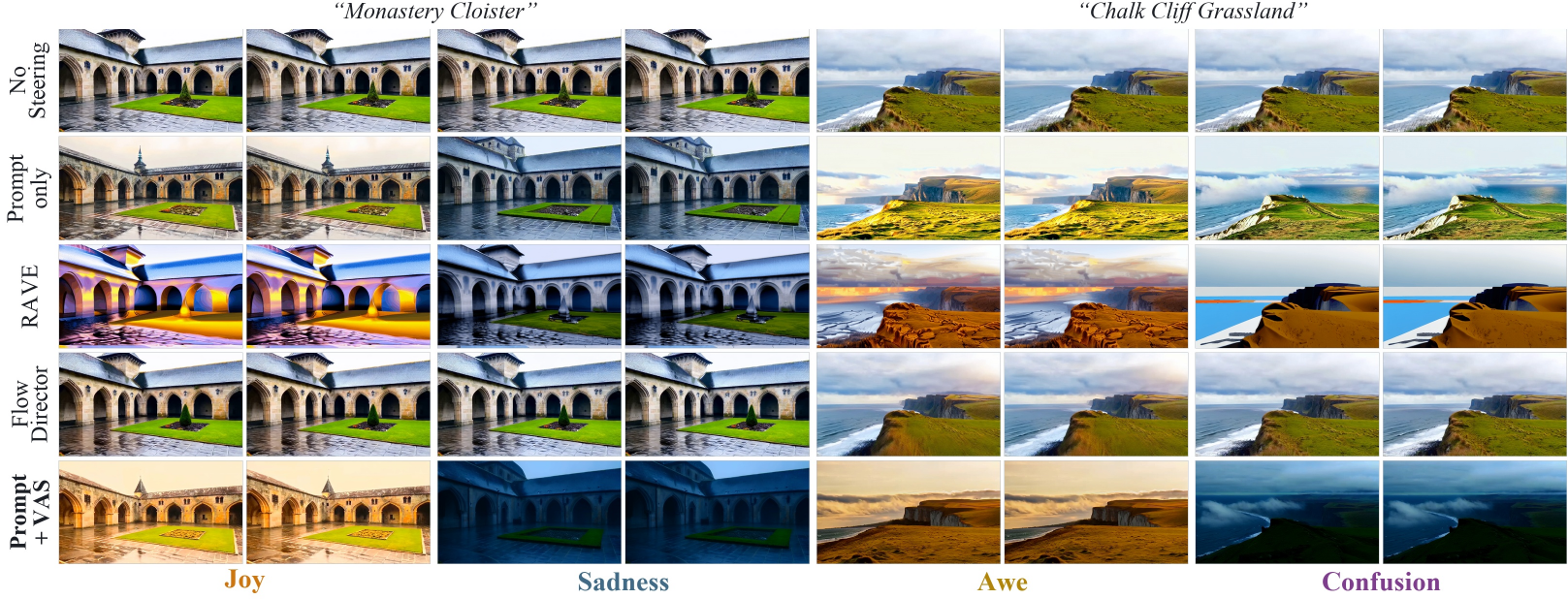}
\caption{\textbf{T2V atmosphere comparison across selected affective categories.} Prompt-only and Prompt+VAS use identical cue-library-composed prompts and matched generation settings.}
\label{fig:qual_vas_t2v}
\end{figure*}

We evaluate VAS on T2V and I2V atmosphere control, SAS on semantic cue control, and TAS on T2V and I2V emotion transitions. The evaluation further covers the complete 27-emotion taxonomy, multiple Video-DiT backbones, and camera-conditioned generation. Matched comparisons share prompts, initializations, and sampling settings; the supplementary material provides complete manifests and implementation details.

\subsection{Experimental Settings}

\paragraph{Evaluation suite and data scale.}
The evaluation covers the complete 27-category Cowen--Keltner taxonomy~\cite{cowen2017self} across diverse indoor and outdoor scenes. For comparisons with external systems, we use a shared subset of emotions spanning valence and arousal; the full taxonomy is used to assess category coverage and structural preservation. VAS vectors are extracted from 649 LayerPano3D panoramas~\cite{layerpano3d}; five geometry-preserving Qwen-Image-Edit variants per panorama yield 3,245 edited observations across the 27-emotion vocabulary. The same pairs produce the frozen Qwen2.5-VL cue library. These data support taxonomy-wide atmosphere control, semantic cue evaluation, temporal transitions, cross-backbone experiments, adjustable steering strength, and camera-conditioned composition.

\paragraph{Models and baselines.}
The main generator is frozen Wan2.2-5B~\cite{wan2025}. VAS is compared with No steering, Prompt-only, RAVE~\cite{rave2024}, FlowDirector~\cite{flowdirector2026}, and an EmoEdit-to-Wan I2V cascade~\cite{emoedit2025}. For atmosphere control, Prompt-only uses the cue-library-composed emotion prompt, and Prompt+VAS adds internal steering to the same prompt. SAS uses Base, Cue, Base+VAS, Cue+VAS, and Cue+VAS+SAS; the last two share cue wording, initialization, seed, VAS
strength, and sampler settings. TAS is compared with Static start, five-pass emotion-field mixing (EFM), and Prompt2Progression~\cite{prompt2prog2025}.

\paragraph{Metrics and implementation.}
CLIP-Emo measures target-emotion alignment; EI denotes its margin over the strongest non-target category; CLIP-Q measures frame quality; EQI combines CLIP-Emo and CLIP-Q; temporal fluctuation (TF) measures adjacent-frame instability; and DINOv2 structural consistency (DINO-SC) measures scene preservation. For SAS, CLIP-Emo measures whole-frame target-affect alignment, while Grounding DINO reports the mean number of detected target cues per sampled frame; full-video cue presence and persistence are reported in the supplementary material. Endpoint alignment, monotonicity, and smoothness evaluate TAS. Unless noted otherwise, Wan outputs use $832\times480$, 49 frames, 20 denoising steps, and CFG $w=5.0$; TAS uses 40 steps. VAS strength is $0.18$ for T2V and $0.15$ for I2V, while SAS uses $\rho=0.20$ with per-frame spatial centering. Remaining settings are in Appendix~\ref{sec:benchmark_scale}.

\subsection{Comparative Results}
\label{sec:comparative}

\begin{figure}[t]
\centering
\includegraphics[width=\columnwidth]{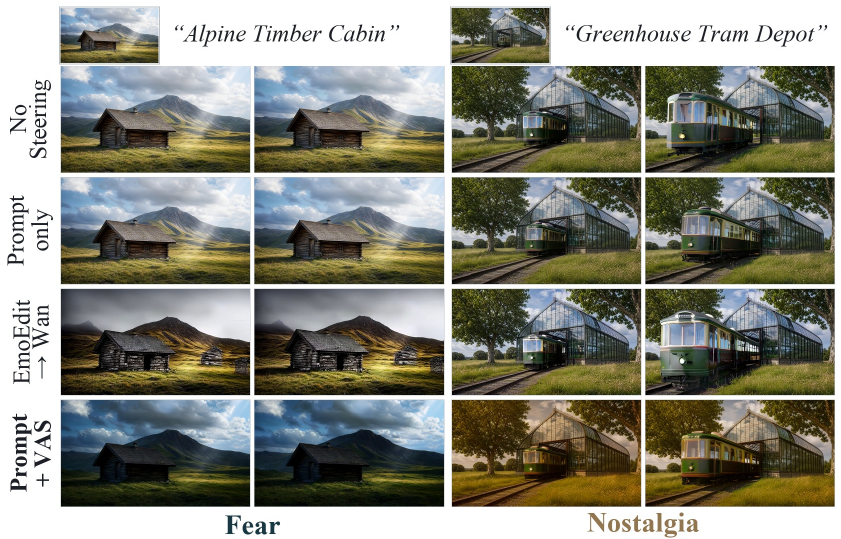}
\caption{\textbf{I2V atmosphere comparison across selected emotion categories.} Given the same reference image and cue-composed prompt, Prompt+VAS strengthens the target atmosphere while preserving the reference scene.}
\label{fig:qual_vas_i2v}
\end{figure}

\paragraph{Single-emotion atmosphere control.}
Table~\ref{tab:single_emotion} shows that Prompt+VAS achieves the strongest aggregate alignment and emotion margin in both generation modes. Relative to Prompt-only, T2V CLIP-Emo improves by 19\% while temporal fluctuation falls by 48\%; I2V alignment also improves under the stronger reference-image constraint. Figures~\ref{fig:qual_vas_t2v} and~\ref{fig:qual_vas_i2v} visualize the corresponding T2V and I2V results.

\begin{table}[t]
\centering\small
\setlength{\tabcolsep}{4pt}
\begin{adjustbox}{max width=\columnwidth}
\begin{tabular}{llcccc}
\toprule
Task & Method & CLIP-Emo $\uparrow$ & EI $\uparrow$ & EQI $\uparrow$ & TF $\downarrow$ \\
\midrule
\multicolumn{6}{l}{\textit{T2V atmosphere control (shared cross-system subset)}} \\
 & No steering & 0.154 & -0.068 & 0.024 & 0.93 \\
 & Prompt-only & 0.168 & -0.049 & 0.029 & 1.22 \\
 & RAVE~\cite{rave2024} & 0.166 & -0.055 & 0.032 & 5.15 \\
 & FlowDirector~\cite{flowdirector2026} & 0.141 & -0.048 & 0.023 & 5.93 \\
 & Prompt+VAS (Ours) & \textbf{0.200} & \textbf{-0.026} & \textbf{0.035} & \textbf{0.63} \\
\midrule
\multicolumn{6}{l}{\textit{I2V atmosphere control (shared cross-system subset)}} \\
 & No steering & 0.159 & -0.063 & 0.025 & 5.82 \\
 & Prompt-only & 0.162 & -0.060 & 0.025 & 12.21 \\
 & EmoEdit$\to$Wan~\cite{emoedit2025} & 0.126 & -0.077 & 0.021 & 64.15 \\
 & Prompt+VAS (Ours) & \textbf{0.171} & \textbf{-0.053} & \textbf{0.027} & 14.19 \\
\bottomrule
\end{tabular}
\end{adjustbox}
\caption{\textbf{Comparative results of single-emotion atmosphere control.} Results use the standardized category subset shared by all compared systems and are macro-averaged with equal category weight. Prompt-only and Prompt+VAS use identical cue-library-composed prompts. External methods use their native control interfaces and serve as reference baselines; TF is in units of $10^{-5}$.}
\label{tab:single_emotion}
\end{table}

% Expected split qualitative assets:
%   fig2/main_qual_t2v_single_emotion_v2.pdf
%   fig2/main_qual_i2v_single_emotion_v2.pdf

\paragraph{Semantic cue control.}
To isolate SAS from cue wording, we compare Cue+VAS with
Cue+VAS+SAS using the same cue-augmented prompt,
initialization, seed, VAS strength, and sampling settings.
In this evaluation, SAS raises CLIP-Emo from
0.155 to 0.212, corresponding to a 36.9\% relative
improvement, and the mean number of detected affect-bearing cues from 2.57 to 3.50 per sampled frame, a relative gain of 36.1\%. More canonical metrics and controlled analyses are in the supplementary material.

\paragraph{Temporal affect evolution.}
TAS achieves the strongest endpoint alignment and monotonicity in both T2V and I2V (Table~\ref{tab:t2v_transition}). In T2V, endpoint alignment improves by 27\% over Static start and monotonicity increases by $5.6\times$; I2V changes from negative monotonicity under Static/EFM to $0.421$. A matched LERP--GSlerp ablation isolates interpolation geometry: GSlerp raises monotonicity from $0.053$ to $0.558$ and smoothness from $0.141$ to $0.439$, supporting great-circle interpolation on the evaluated transitions.

\begin{table}[t]
\centering\small
\setlength{\tabcolsep}{4pt}
\begin{adjustbox}{max width=\columnwidth}
\begin{tabular}{llcc}
\toprule
Task & Method & End Align. $\uparrow$ & Mono. $\uparrow$ \\
\midrule
\multicolumn{4}{l}{\textit{T2V transition}} \\
 & Static start & 0.163 & 0.141 \\
 & EFM (5-pass LERP) & 0.152 & 0.687 \\
 & Prompt2Progression~\cite{prompt2prog2025} & 0.143 & 0.435 \\
 & TAS (Ours) & \textbf{0.207} & \textbf{0.788} \\
\midrule
\multicolumn{4}{l}{\textit{I2V transition}} \\
 & Static start & 0.161 & -0.507 \\
 & EFM (5-pass LERP) & 0.177 & -0.325 \\
 & TAS (Ours) & \textbf{0.185} & \textbf{0.421} \\
\bottomrule
\end{tabular}
\end{adjustbox}
\caption{\textbf{Comparative results of temporal affect transitions.} TAS interpolates VAS-conditioned endpoint residual fields using a frame- and denoising-dependent coordinate field and GSlerp. The T2V and I2V evaluations contain 27 and 18 generated videos, respectively.}
\label{tab:t2v_transition}
\end{table}

\subsection{Component Analysis}
\label{sec:ablation}

\begin{figure}[t]
\centering
\includegraphics[width=\columnwidth]{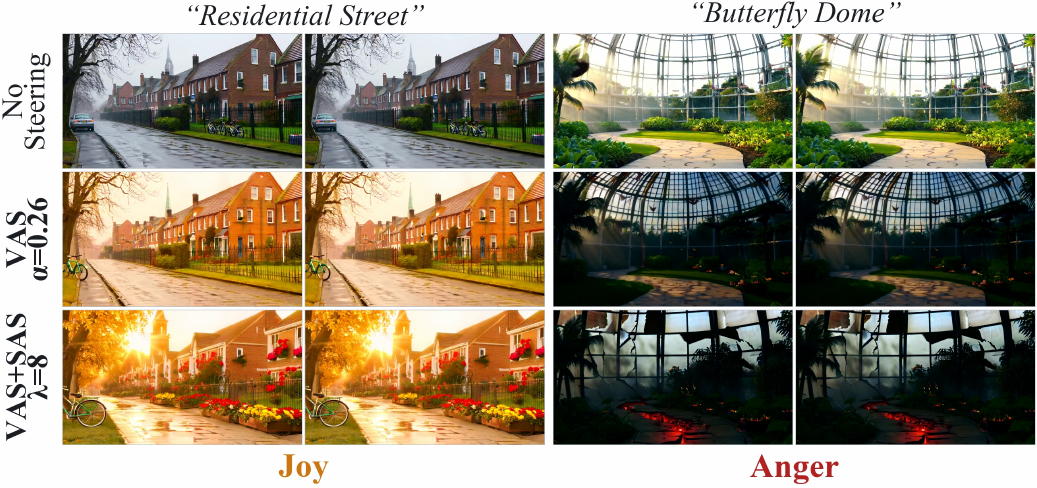}
\caption{\textbf{Qualitative ablations of VAS and SAS.} VAS establishes the global visual atmosphere through block-level steering, whereas SAS modulates affect-bearing semantic cues through the sparse prompt-residual branch. Their combination provides complementary control. }
\label{fig:ablation_vas_sas}
\end{figure}

\begin{figure}[t]
\centering
\includegraphics[width=\columnwidth]{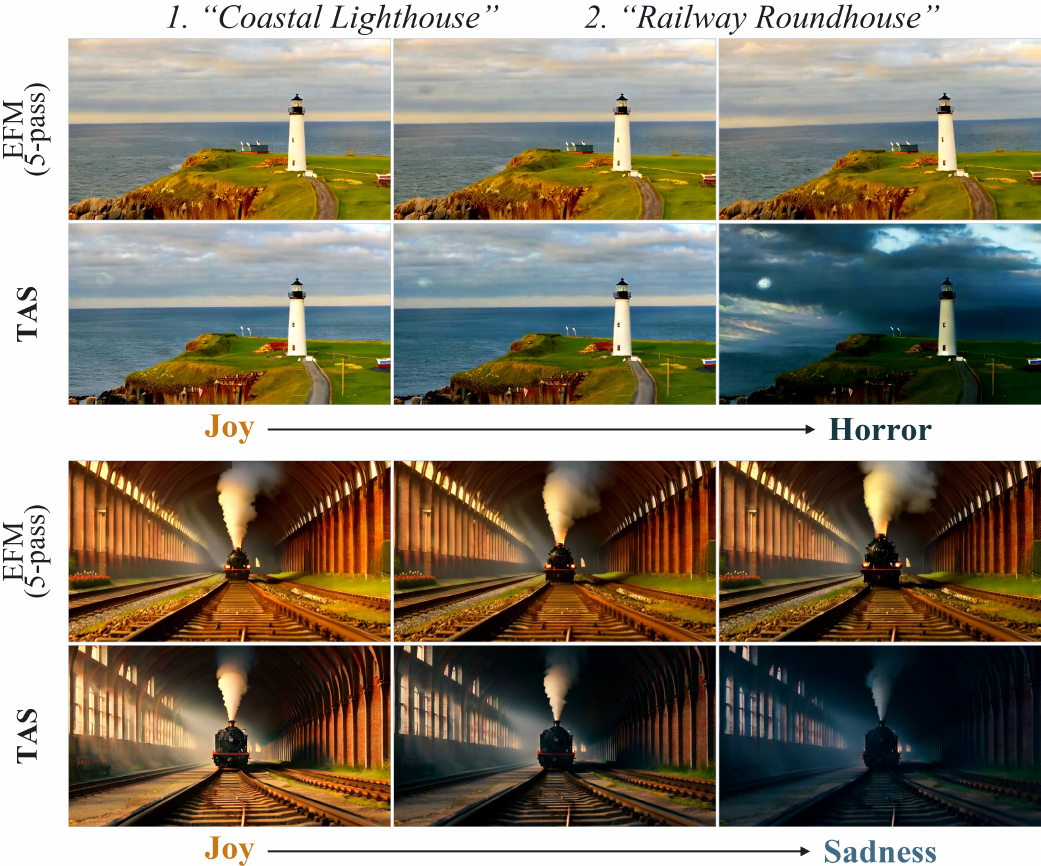}
\caption{\textbf{Qualitative ablation of TAS.} Compared with a static start field and linear endpoint mixing, TAS interpolates endpoint residual directions across video frames to produce a clearer affective progression.}
\label{fig:ablation_tas}
\end{figure}

Because the three components target different aspects of affect,
Table~\ref{tab:ablation} reports targeted comparisons for each
control target. Complete metric sets and controlled ablations of
hook placement, SAS residual shaping, endpoint construction, and
interpolation geometry are provided in the supplementary material.
As a result, VAS improves global atmosphere beyond emotion prompting, while the cue-matched SAS comparison strengthens whole-frame target-affect alignment and increases the realization of detected affect-bearing cues. Additional analyses in the supplementary material further examine SAS residual routing. Figure~\ref{fig:ablation_vas_sas}
illustrates these complementary effects. TAS outperforms all
evaluated transition baselines, with the corresponding temporal
behavior shown in Figure~\ref{fig:ablation_tas}. Together, these
results support separate control of atmosphere, semantic cues,
and temporal progression.

\begin{table}[t]
\centering
\small
\setlength{\tabcolsep}{3.5pt}
\begin{adjustbox}{max width=\columnwidth}
\begin{tabular}{@{}llll@{}}
\toprule
Component & Control target & Evaluation comparison & Target metric \\
\midrule
VAS
& Atmosphere
& Prompt-only $\rightarrow$ Prompt+VAS
& CLIP-Emo: $0.168 \rightarrow \mathbf{0.200}$ \\
SAS
& Semantic cues
& Cue+VAS $\rightarrow$ Cue+VAS+SAS
& \begin{tabular}[c]{@{}l@{}}
  CLIP-Emo: $0.155 \rightarrow \mathbf{0.212}$ \\
  Cues/frame: $2.57 \rightarrow \mathbf{3.50}$
  \end{tabular} \\
TAS
& Progression
& EFM (5-pass LERP) $\rightarrow$ TAS
& Monotonicity: $0.687 \rightarrow \mathbf{0.788}$ \\
\bottomrule
\end{tabular}
\end{adjustbox}
\caption{\textbf{Targeted evaluation of the three control components.}
Each row reports the effect of adding the corresponding
operator to its preceding control configuration, using metrics aligned
with the intended control target. Complete canonical metrics, matched
ablations, and sensitivity analyses are provided in the supplementary
material.}
\label{tab:ablation}
\end{table}

% Expected split ablation assets:
%   fig2/main_ablation_vas_sas.pdf
%   fig2/main_ablation_tas.pdf

\paragraph{Breadth, portability, and composability.}
VAS exposes a continuously adjustable steering-strength parameter and can be instantiated on CogVideoX and VMem using backbone-specific probes, steering vectors, and compatible hooks. The camera-conditioned path composes viewpoint motion with scheduled atmosphere steering. Across the complete 27-emotion benchmark, aggregate T2V and I2V results show broad category coverage while DINO-SC remains high ($\geq0.88$). Full category distributions, alignment--preservation analysis, architectural portability, and transition trajectories appear in the supplementary material.

\section{Discussion and Conclusion}

We have presented EmoWorld, which decomposes emotional video control into atmosphere, semantic cues, and temporal progression within a frozen Video DiT. Paired neutral and edited panoramas provide feature- and language-space affect representations, while VAS, SAS, and TAS act on hidden states, prediction residuals, and frame-wise velocity fields. Across 27 emotion categories, the framework supports T2V and I2V generation, backbone-specific instantiations, and camera-conditioned trajectories while preserving recognizable scene structure.

Beyond the individual operators, EmoWorld embodies a support-matching principle for controllable generation: each affective factor is represented at the spatial and temporal scale at which it acts. VAS uses layer-wise directions for scene-wide and persistent appearance; SAS removes spatially constant prompt effects and retains sparse residual coordinates for localized cues; and TAS defines a continuous path over denoising and frame coordinates for evolving affect. This perspective explains why the three operators can be scaled and evaluated separately, and suggests a general recipe for factorizing composite video controls into global, local, and temporal components rather than forcing them to compete within a single conditioning channel.

\paragraph{Limitations.}
EmoWorld mainly controls environmental affect rather than facial expression, character action, or narrative causality. The method also requires offline paired edits, backbone-specific feature extraction, and multiple forward passes. Future work should address human-centered evaluation, region-aware control, and efficient joint inference.

\bibliography{aaai2027}
\clearpage

\appendix
\section*{Supplementary Material}

This supplementary material provides the implementation and
evaluation details supporting the main paper. We first summarize
the evaluation suite, sampling protocols, metrics, and operator
configurations, followed by additional details on VAS, SAS, TAS,
camera-conditioned composition, and computational cost. We then
report a canonical SAS test-set evaluation, a cue-matched
residual-routing control, the complete TAS endpoint--interpolation
factorial test set, controlled sensitivity studies, focused human
perceptual validation, paired affective preparation, taxonomy-wide
and cross-backbone results, and additional qualitative examples.

\section{Evaluation Suite and Protocols}
\label{sec:benchmark_scale}

The evaluation spans emotion category, generation mode, operator, transition profile, model architecture, and camera control. The complete 27-emotion taxonomy provides category-wide coverage, while a standardized shared subset supports comparisons with external baselines. Table~\ref{tab:benchmark_scale} summarizes the complementary tracks. Exact video identities, prompts, initializations, active operators, and sampling configurations are recorded in the accompanying manifests.

\begin{table*}[t]
\centering\small
\begin{adjustbox}{max width=\linewidth}
\begin{tabular}{llll}
\toprule
Evaluation track & Coverage & Controlled factors & Reported metrics \\
\midrule
Paired affective preparation & 27 emotions; 649 panoramas; 3,245 edits & scene identity, geometry, viewpoint & steering vectors and cue library \\
Taxonomy-wide atmosphere control & complete 27-emotion taxonomy; T2V and I2V & category, strength, generation mode & alignment, margin, DINO-SC \\
Standardized cross-system comparison & shared valence--arousal subset & method under common prompts and settings & CLIP-Emo, EI, EQI, TF \\
Semantic cue control & multiple scenes, methods, and cue families & VAS and SAS residual design & cue presence, persistence, alignment \\
Temporal affect progression & multi-pair T2V/I2V benchmark; complete $2\times2$ endpoint--interpolation test set & endpoint construction and interpolation geometry & endpoint alignment, monotonicity, smoothness, path linearity \\
Focused human perceptual validation & 54 participants; 1944 judgments & matched pairwise comparisons for VAS, SAS, and TAS & atmosphere, cue realization, transition clarity and smoothness \\
Backbone portability and camera control & Wan2.2, CogVideoX, VMem; camera paths & backbone and viewpoint control & quantitative and qualitative comparisons \\
\bottomrule
\end{tabular}
\end{adjustbox}
\caption{\textbf{EmoWorld evaluation suite.} The tracks jointly test category breadth, method-level gains, semantic-cue control, temporal progression, structural preservation, architectural portability, and composability.}
\label{tab:benchmark_scale}
\end{table*}

Unless otherwise stated, Wan outputs use $832\times480$ resolution, 49 frames, 20 denoising steps, and CFG $w=5.0$. The T2V VAS comparison uses strength $0.18$; I2V caps it at $0.15$. SAS uses $\lambda_{\mathrm{sem}}=5.0$, retention ratio $\rho=0.20$, per-frame spatial centering, warmup endpoint $a=0.05$, fade-start $b=0.65$, and VAS strength $0.18$ where enabled. TAS uses 40 steps, VAS strength $0.22$, and the transition settings listed in Section~\ref{sec:tas_mechanics}.

\section{Additional Technical Details}

\subsection{Flow-Matching and Classifier-Free Guidance}

Let $\mathbf{v}_{\theta}(\mathbf{x}_t,t,\mathbf{c})$ denote the frozen video DiT's velocity prediction at latent $\mathbf{x}_t$, denoising time $t$, and condition $\mathbf{c}$. Standard classifier-free guidance (CFG) is
\begin{equation}
    \mathbf{v}_{\mathrm{CFG}}
    =
    \mathbf{v}_{\varnothing}
    + w(\mathbf{v}_{c}-\mathbf{v}_{\varnothing}),
    \label{eq:supp_cfg}
\end{equation}
where $w$ is the guidance scale. VAS modifies a conditional forward pass through feature hooks; SAS and TAS assemble multiple predictions before the scheduler step. The generator weights remain unchanged in all cases.

\subsection{Metric Protocol}
\label{sec:metric_protocol}

The reported tables use a manifest-driven evaluator with normalized CLIP ViT-L/14 embeddings. Model revisions, source-video identities, prompts, and table manifests are fixed and recorded in the accompanying protocol files. Let $F_V$ be the full decoded frame count and $F_C=\min(81,F_V)$; here $F_V=F_C=49$. Let $\mathbf{z}_f$ be the normalized CLIP embedding of frame $f$, $\mathbf{t}_e$ the normalized embedding of the fixed target-emotion text, and $\mathbf{t}_Q$ the embedding of the exact quality text ``A high quality, detailed, clear, well-lit video frame.'' Over the first $F_C$ frames in temporal order,
\begin{equation}
\begin{split}
\mathrm{CLIP\mbox{-}Emo}
&=\frac{1}{F_C}\sum_{f=1}^{F_C} \mathbf{z}_f^\top\mathbf{t}_e,\\
\mathrm{EI}
&=\frac{1}{F_C}\sum_{f=1}^{F_C}\left(\mathbf{z}_f^\top\mathbf{t}_e-
\max_{e'\neq e}\mathbf{z}_f^\top\mathbf{t}_{e'}\right),\\
\mathrm{CLIP\mbox{-}Q}
&=\frac{1}{F_C}\sum_{f=1}^{F_C} \mathbf{z}_f^\top\mathbf{t}_Q,\\
\mathrm{EQI}
&=\mathrm{CLIP\mbox{-}Emo}\cdot\mathrm{CLIP\mbox{-}Q}.
\end{split}
\label{eq:metric_clip}
\end{equation}
EI is a CLIP emotion margin rather than a direct psychophysical measure of intensity; the competitor is the highest-scoring non-target text in the fixed 27-emotion set. EQI is computed per video before rounding, and both component scores remain visible because their product can hide which factor changed. The exact 27 texts are recorded with the source protocol and are distinct from the generation prompts used by SAS.

For grayscale Canny edge maps $C_f$ computed with thresholds 50 and 150, write $J(A,B)=1$ when $A\cup B=\varnothing$ and $J(A,B)=|A\cap B|/|A\cup B|$ otherwise, matching the evaluator's empty-edge convention. For normalized grayscale frames $G_f$, the temporal proxies are
\begin{equation}
\begin{split}
\mathrm{Edge\mbox{-}SC}
&=\frac{1}{F_C-1}\sum_{f=1}^{F_C-1}J(C_f,C_{f+1}),\\
d_f&=\operatorname{mean}|G_{f+1}-G_f|,
\quad 1\leq f<F_V,\\
\mathrm{TF}&=\operatorname{Var}_{1\leq f<F_V}(d_f).
\end{split}
\label{eq:metric_temporal}
\end{equation}
Edge-SC uses up to the first 81 frames; TF decodes the full video. They measure adjacent-frame stability and can respond to camera motion or intended semantic change. Neither measures retention relative to an input image or neutral reference, which is assessed separately.

\paragraph{Task-specific metric conventions.}
All task-specific scores are computed per video before aggregation. DINO-SC compares normalized DINOv2 representations of generated frames with the matched structural reference specified by the corresponding T2V or I2V track. Grounding DINO cue presence is the fraction of decoded frames in which the requested affect-bearing concept is detected under a fixed detector configuration; cue persistence summarizes the temporal consistency of those detections. TAS endpoint alignment (EES), monotonicity, and smoothness are derived from frame-wise start- and end-emotion score trajectories. For the complete factorial test set, Path Linearity measures agreement between the observed affect trajectory and the prescribed transition path (higher is better). The model revisions, detector thresholds, endpoint windows, trajectory normalization, and aggregation settings are fixed in the evaluator and recorded in the accompanying protocol files. When uncertainty is reported, the resampling unit is the matched scene--emotion--seed group rather than an individual frame, and paired comparisons reuse the same bootstrap draws.

\subsection{Evaluated VAS Hook Path}

In the evaluated Wan path, self-attention input hooks are registered at blocks $\{0,5,10,15,20,25,29\}$, and cross-attention output hooks are additionally registered at blocks $\{15,20,25,29\}$. A per-emotion unit steering vector is broadcast over the video tokens at each feature hook. Self-attention coefficients use the user atmosphere strength plus depth, denoising-step, and frame schedules. The late cross-attention hook uses a separate layer- and emotion-specific coefficient and does not inherit that same step/frame schedule. Injection is enabled for the conditional VAS pass and disabled for the matched base, emotion-prompt, and unconditional SAS passes.

These hook routes are implementation sites rather than separate semantic, style, and intensity channels. Equation~\ref{eq:vas_injection} specifies the applied steering vector and gain without assigning unsupported physical meanings to individual hooks.

\subsection{SAS Tensor Convention and Cost}
\label{sec:sas_tensor_cost}

After selecting the batch item, the Wan velocity tensor passed to the spatial projector has shape $[C,F,H,W]$. The projector averages only the two spatial axes, leaving channel and latent-frame indices independent. The sparsifier computes a single absolute-magnitude quantile over all scalar entries of the projected $C\times F\times H\times W$ tensor and retains entries at or above the threshold. It is therefore coordinate-wise and channel-sensitive; no connected-component, token, or temporal grouping is applied.

VAS+SAS uses four predictions of the same latent at each denoising step: base prompt without VAS, cue-augmented prompt without VAS, base prompt with VAS, and unconditional prompt. A standard CFG baseline uses two predictions. SAS-only can omit the VAS pass, while VAS-only uses the standard unconditional and VAS-conditional pair. Warmup, fade-start, retention ratio $\rho$, semantic strength, atmosphere strength, retrieved cues, and the exact composed prompt pair are recorded for every reported output.

The main experiments use a warmup--hold--fade schedule. With normalized step $\xi=i/N$, warmup endpoint $a$, and fade-start $b$,
\begin{equation}
 s(\xi)=
 \begin{cases}
 \xi/a, & 0\leq\xi<a,\\
 1, & a\leq\xi\leq b,\\
 (1-\xi)/(1-b), & b<\xi\leq1.
 \end{cases}
 \label{eq:sas_schedule_supp}
\end{equation}
The runtime parameter named \texttt{semantic\_stop} therefore starts a linear fade rather than an immediate stop. Boundary cases are handled explicitly in code.

\subsection{TAS Transition Mechanics}
\label{sec:tas_mechanics}

TAS uses three predictions per denoising step: (1) a start-emotion conditional pass with the start VAS steering vector injected, (2) an end-emotion conditional pass with the end VAS steering vector injected, and (3) fixed negative conditioning. Because each prediction contains all latent frames, the two emotion residuals define endpoint vector fields over denoising step $q$ and frame index $f$. The scalar coordinate field $\beta_{q,f}$ is initialized from the requested narrative profile, relaxed across denoising steps, and smoothed over neighboring frames. GSlerp in Eq.~\ref{eq:tas_slerp} is then applied after flattening one complete latent frame over channels and spatial coordinates. The residual direction follows great-circle interpolation when nondegenerate, while residual norm is interpolated linearly between endpoint norms. Near-zero, nearly parallel, and nearly antipodal cases use numerical fallbacks. The evaluated TAS path uses VAS-conditioned endpoint fields, whereas EFM constructs endpoints from prompt differences without feature steering. This comparison separates both the endpoint construction and the geometry used to interpolate the affective velocity.

The evaluated TAS configuration uses: 3 scenes (cathedral\_nave, old\_town\_square, residential\_street), 3 emotion pairs (joy$\to$sadness, joy$\to$horror, calmness$\to$horror), 3 seeds (42, 123, 629), 49 frames, 40 denoising steps, CFG $w=5.0$, VAS strength $\alpha=0.22$, stretched-sigmoid schedule with $\beta_{\min}=0.1$, $\beta_{\max}=0.9$, and sharpness $3.0$. I2V uses the same settings with middle-to-end frame transition ($t_{\text{mid}}=F_{\text{latent}}/2$). T2V yields 27 videos and I2V yields 18 videos. SAS and TAS are evaluated independently in these experiments.

\subsection{Camera-Specific Composition Path}

We instantiate camera-aware affective control using the
\texttt{Wan2.2-Fun-5B-Control-Camera} backbone, a
camera-conditioned variant of Wan2.2-TI2V-5B that accepts
prescribed viewpoint trajectories.
The camera-control pathway determines viewpoint motion, while
EmoWorld applies scheduled VAS directions to the Video-DiT hidden
states to control scene-wide affect. Because these controls enter
through distinct pathways, they can be composed at inference
without updating the generator parameters.

\subsection{Computing Environment and Inference Cost}
\label{sec:compute_cost}

All experiments were conducted on a single NVIDIA RTX 6000 Ada
Generation GPU with 48\,GB of memory. Table~\ref{tab:inference_cost} reports the measured inference cost
of the principal configurations. We use the same T2V prompt and
random seed for all measurements. Each configuration is warmed up
once and then measured over three repeated timing runs. Runtime
covers the denoising loop and excludes model loading, text
encoding, VAE decoding, and video saving. CUDA synchronization is
applied before and after timing, and peak memory is measured using
PyTorch CUDA memory statistics.

\begin{table*}[t]
\centering
\small
\setlength{\tabcolsep}{5pt}
\begin{tabular}{lccccccc}
\toprule
Configuration
& Steps
& DiT calls / step
& Total calls
& Runtime (s)
& Peak alloc. (GB)
& Peak reserv. (GB)
& Relative time \\
\midrule
Base / Prompt-only
& 20 & 3 & 60
& $47.99 \pm 0.22$
& 23.18 & 23.36 & $1.000\times$ \\

Prompt+VAS
& 20 & 3 & 60
& $49.68 \pm 2.18$
& 23.18 & 23.36 & $1.035\times$ \\

Cue+VAS+SAS
& 20 & 4 & 80
& $63.44 \pm 0.18$
& 23.19 & 23.36 & $1.322\times$ \\

TAS
& 40 & 3 & 120
& $94.78 \pm 0.28$
& 23.18 & 23.36 & $1.975\times$ \\
\bottomrule
\end{tabular}
\caption{\textbf{Measured inference cost on one NVIDIA RTX 6000
Ada Generation GPU.}
All measurements use Wan2.2-TI2V-5B at $832\times480$ resolution
with 49 output frames and CFG scale 5.0. Values are the mean and
standard deviation over three timing runs after one warm-up run.
Runtime covers denoising only and excludes model initialization and all pre- and post-processing.}
\label{tab:inference_cost}
\end{table*}

In the current unified implementation, Base and Prompt+VAS both
execute three Video-DiT calls per denoising step. The third
base-conditioned branch is retained even when its steering strength is zero and therefore represents a removable implementation overhead rather than an intrinsic requirement of the base method. VAS introduces only a 3.5\% runtime increase, whereas SAS adds one cue-conditioned call and increases runtime by 32.2\%. TAS evaluates the start-emotion, end-emotion, and unconditional branches separately for 40 denoising steps, requiring 120 calls in total and $1.975\times$ the runtime of Base/Prompt-only. Peak allocated memory remains below 23.2\,GB across all configurations.

\section{Additional Ablation Details}

\begin{figure*}[t]
\centering
\includegraphics[width=0.9\textwidth]{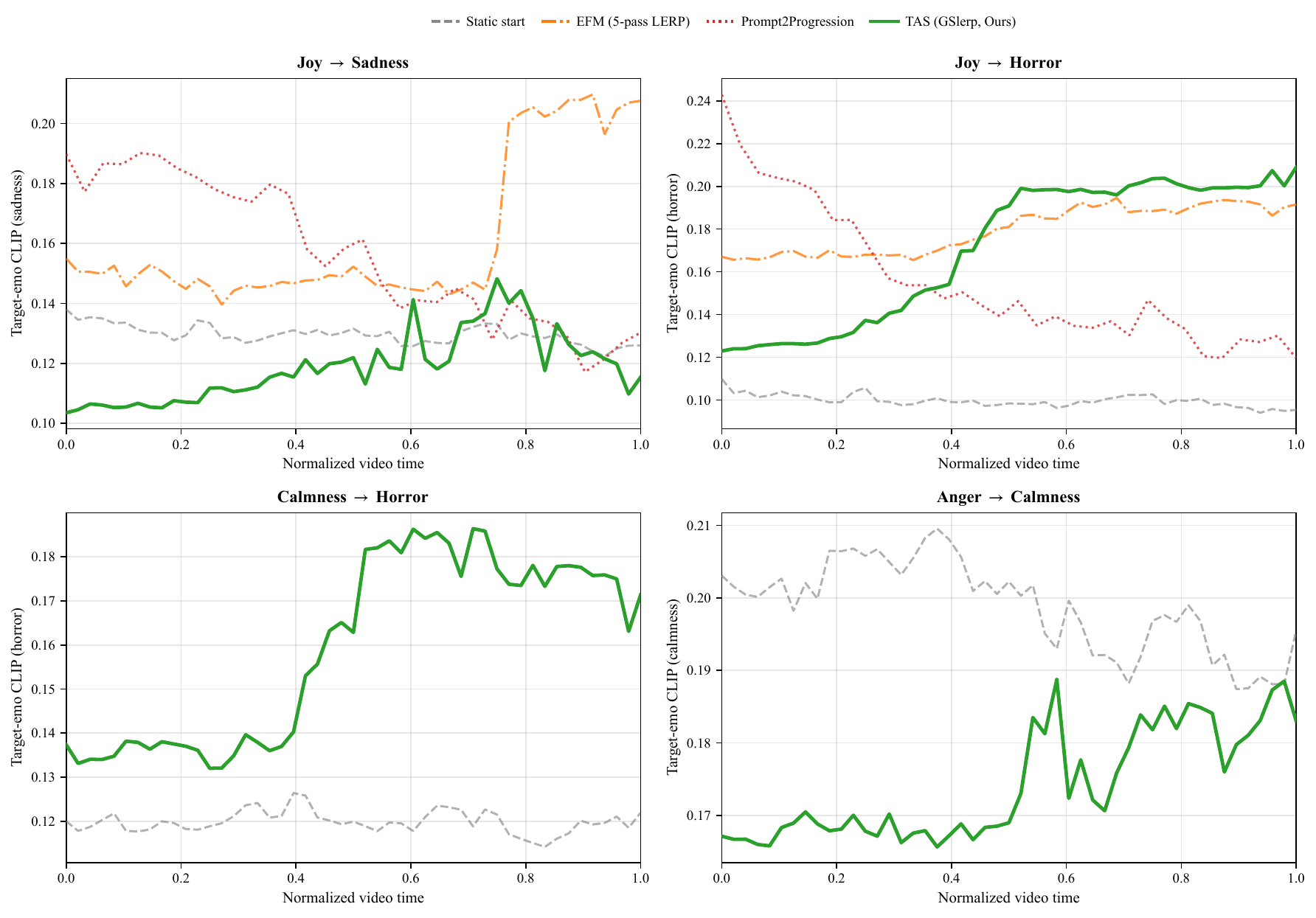}
\caption{\textbf{Frame-wise affect-transition trajectories.} Target-emotion CLIP scores across four representative transitions. TAS shows clear directional progression on the more challenging joy-to-horror and calmness-to-horror transitions, while trajectory shape and endpoint behavior remain
transition-dependent. Baselines are shown where matched results are available; aggregate endpoint and monotonicity comparisons are reported in Table~\ref{tab:t2v_transition}, and the matched interpolation ablation is reported in Table~\ref{tab:tas_ablation}.}
\label{fig:tas_trajectory}
\end{figure*}

\paragraph{Multi-pair TAS interpolation comparison.}
Table~\ref{tab:tas_ablation} compares linear chord interpolation (LERP) with great-circle interpolation (GSlerp) on three transition pairs. GSlerp yields higher endpoint alignment on all three pairs and substantially higher macro monotonicity ($0.558$ vs.\ $0.053$) and smoothness ($0.439$ vs.\ $0.141$). This matched ablation supports the TAS design principle: interpolating residual direction on the unit hypersphere avoids the norm shrinkage and cancellation that can arise along the linear chord. The resulting TAS trajectories are visualized separately in Figure~\ref{fig:tas_trajectory}.

\begin{table}[t]
\centering\small
\setlength{\tabcolsep}{4pt}
\begin{adjustbox}{max width=\columnwidth}
\begin{tabular}{llccccc}
\toprule
Pair & Interp. & Start $\uparrow$ & Mid $\uparrow$ & End $\uparrow$ & Mono $\uparrow$ & Smooth $\uparrow$ \\
\midrule
Joy $\to$ Sadness & LERP & 0.139 & 0.181 & 0.180 & 0.624 & 0.439 \\
 & GSlerp (Ours) & 0.137 & 0.193 & 0.197 & 0.201 & 0.339 \\
\midrule
Joy $\to$ Horror & LERP & 0.143 & 0.124 & 0.123 & -0.589 & 0.101 \\
 & GSlerp (Ours) & 0.143 & 0.130 & 0.130 & 0.650 & 0.421 \\
\midrule
Calmness $\to$ Horror & LERP & 0.146 & 0.109 & 0.109 & 0.123 & -0.116 \\
 & GSlerp (Ours) & 0.169 & 0.129 & 0.130 & 0.823 & 0.558 \\
\midrule
\textbf{Macro Avg} & \textbf{LERP} & \textbf{0.143} & \textbf{0.138} & \textbf{0.137} & \textbf{0.053} & \textbf{0.141} \\
 & \textbf{GSlerp (Ours)} & \textbf{0.150} & \textbf{0.151} & \textbf{0.153} & \textbf{0.558} & \textbf{0.439} \\
\bottomrule
\end{tabular}
\end{adjustbox}
\caption{\textbf{TAS interpolation ablation.} On macro average, GSlerp improves endpoint alignment, monotonicity, and smoothness over linear interpolation, supporting great-circle interpolation of affective residual directions.}
\label{tab:tas_ablation}
\end{table}

\paragraph{Complete TAS endpoint--interpolation test set.}
We further evaluate the complete TAS factorial test set using a
$2\times2$ design that crosses endpoint construction with
interpolation geometry. The four test conditions exhaust all
combinations of prompt-difference or VAS-conditioned endpoints
with LERP or GSlerp under the same joy-to-horror transition
protocol. All conditions share the prompt, initialization, seed,
sampler, and temporal profile. Prompt-difference endpoints use
zero feature-steering strength, whereas VAS-conditioned endpoints
use strength $0.22$. This factorial protocol uses the updated
\texttt{compute\_metrics\_v3.py} evaluator and is distinct from the
multi-pair interpolation ablation in Table~\ref{tab:tas_ablation};
the absolute values of metrics with different definitions are
therefore not directly interchangeable.

\begin{table}[t]
\centering
\small
\setlength{\tabcolsep}{4.5pt}
\begin{adjustbox}{max width=\columnwidth}
\begin{tabular}{lllccc}
\toprule
ID & Endpoint & Interp. &
EES $\uparrow$ &
Path Linear. $\uparrow$ &
Mono. $\uparrow$ \\
\midrule
P-L & Prompt difference & LERP
& 0.1633 & 0.3993 & 0.5521 \\
P-G & Prompt difference & GSlerp
& \textbf{0.1914} & 0.8274 & 0.5104 \\
V-L & VAS conditioned & LERP
& 0.1729 & 0.9269 & 0.5521 \\
V-G & VAS conditioned & GSlerp
& 0.1796 & \textbf{0.9582} & \textbf{0.5729} \\
\bottomrule
\end{tabular}
\end{adjustbox}
\caption{\textbf{Endpoint--interpolation factorial analysis on the complete TAS test set.}
The $2\times2$ design isolates endpoint construction and interpolation geometry under matched generation settings. EES measures arrival at the target emotion, Path Linearity measures agreement with the prescribed transition path, and monotonicity measures consistent progression toward the endpoint.}
\label{tab:tas_2x2}
\end{table}

The factorial analysis reveals complementary effects of endpoint
construction and interpolation geometry. With prompt-difference
endpoints, replacing LERP with GSlerp improves EES from $0.1633$
to $0.1914$ and Path Linearity from $0.3993$ to $0.8274$, while
monotonicity changes from $0.5521$ to $0.5104$. With
VAS-conditioned endpoints, GSlerp improves all three reported
metrics: EES increases from $0.1729$ to $0.1796$, Path Linearity
from $0.9269$ to $0.9582$, and monotonicity from $0.5521$ to
$0.5729$.

Among the four configurations, P-G achieves the strongest endpoint
alignment, whereas V-G achieves the highest Path Linearity and
monotonicity. Thus, GSlerp improves endpoint arrival and path
linearity under both endpoint constructions, while its effect on
monotonicity depends on the endpoint field. This interaction
supports jointly considering endpoint construction and interpolation
geometry when designing affective transition controls.

\paragraph{Additional transition trajectories.}
Figure~\ref{fig:tas_trajectory} reports frame-wise target-emotion trajectories across four representative transition pairs. The trajectories complement the aggregate metrics by exposing transition-dependent temporal behavior throughout the generated clip.

\paragraph{Canonical SAS test-set evaluation.}
We further evaluate the complete semantic-control stage using
the paper-official evaluator. The SAS test set contains two
scene--emotion settings, residential-street joy and
butterfly-dome anger, with seven condition-specific videos in
total. All metrics are computed over the complete 49-frame
videos using normalized CLIP ViT-L/14 embeddings and the
fixed 27-emotion text set described in Section~B. Since
CLIP-Emo and detected cues per frame are already reported
in the main paper, Table~\ref{tab:sas_canonical} reports the
complementary affect-margin, quality, structural, and temporal
metrics.

\begin{table}[t]
\centering
\small
\setlength{\tabcolsep}{3.2pt}
\begin{adjustbox}{max width=\columnwidth}
\begin{tabular}{lccccc}
\toprule
Method
& EI $\uparrow$
& Edge-SC $\uparrow$
& CLIP-Q $\uparrow$
& TC $\uparrow$
& Emo-Smooth $\downarrow$ \\
\midrule
VAS
& -0.0875
& \textbf{0.8788}
& 0.1786
& \textbf{0.9992}
& \textbf{0.0019} \\

Cue+VAS+SAS
& \textbf{-0.0369}
& 0.8336
& \textbf{0.1791}
& 0.9990
& 0.0023 \\
\bottomrule
\end{tabular}
\end{adjustbox}
\caption{\textbf{Complementary metrics on the SAS test set.}
Results are macro-averaged over the residential-street joy and
butterfly-dome anger settings. EI measures the target-emotion
margin over the strongest non-target emotion; Edge-SC measures
adjacent-frame edge consistency; TC is adjacent-frame CLIP
similarity; and Emo-Smooth measures second-order variation of
the target-emotion trajectory. CLIP-Emo and detected-cue results
are reported in the main paper.}
\label{tab:sas_canonical}
\end{table}

Across the test set, semantic control improves EI from
$-0.0875$ to $-0.0369$, substantially moving the generated
videos toward the target emotion relative to the strongest
competing category. CLIP-Q remains essentially unchanged,
and adjacent-frame CLIP consistency remains approximately
$0.999$, indicating that the additional semantic content does
not cause a meaningful loss of perceptual temporal consistency.
TF is below the displayed precision for both evaluated
configurations.

Edge-SC decreases from $0.8788$ to $0.8336$. This change
is concentrated in the butterfly-dome anger example, where
semantic control introduces additional structures such as fire
and broken-glass-like details. The result reflects a trade-off
between preserving the original edge layout and realizing new
affect-bearing semantic content, rather than temporal
instability.

\paragraph{Cue-matched residual-routing control.}
For the residential-street setting, we additionally compare
Cue+VAS with Cue+VAS+SAS using the same cue-augmented
prompt and generation settings. EI improves from $-0.0718$
to $-0.0322$, while CLIP-Q increases from $0.1746$ to
$0.1788$. Edge-SC changes only modestly from $0.8554$ to
$0.8493$, and TC increases from $0.9991$ to $0.9993$.
This controlled comparison supports the contribution of SAS
residual routing beyond cue-enriched prompting alone.

\paragraph{Cue-level semantic analysis.}
On the matched cue-groundable subset, the observed mean cue presence is $\SASMatchedCueFull\%$ for Cue+VAS+SAS and $\SASMatchedCueControl\%$ for Cue+VAS, a difference of $\SASMatchedCueGain$ percentage points. A paired hierarchical bootstrap over the matched evaluation groups gives a 95\% interval of $\SASMatchedCueGainCI$ percentage points, and cue persistence changes by $\SASMatchedPersistenceGain$ percentage points. Because the interval includes zero, we treat this detector-based gain as complementary to the stronger whole-frame alignment and cues-per-frame results reported in the main paper.

\paragraph{SAS sensitivity to residual shaping.}
Table~\ref{tab:sas_ablation} varies per-frame spatial centering and magnitude-based coordinate sparsification while holding the VAS branch fixed. No single configuration dominates every whole-frame metric, indicating that centering and retention primarily shape how the semantic residual is distributed rather than guaranteeing a uniform increase in global emotion similarity.

\begin{table}[t]
\centering\small
\setlength{\tabcolsep}{3pt}
\begin{adjustbox}{max width=\columnwidth}
\begin{tabular}{lcccccc}
\toprule
Config & Joy & Sad. & Horr. & Calm. & EI & Edge-SC \\
 & \multicolumn{4}{c}{CLIP-Emo $\uparrow$} & $\uparrow$ & $\uparrow$ \\
\cmidrule(lr){2-5} \cmidrule(lr){6-6} \cmidrule(lr){7-7}
\midrule
Raw prompt residual$^{\dagger}$ & 0.190 & 0.236 & 0.180 & 0.147 & -0.039 & 0.757 \\
VAS+SAS (no center, $\rho{=}0.20$) & 0.186 & 0.210 & 0.185 & 0.145 & -0.031 & 0.699 \\
VAS+SAS (center, dense) & 0.180 & 0.233 & 0.176 & 0.145 & -0.047 & 0.713 \\
VAS+SAS (center, $\rho{=}0.10$) & 0.164 & 0.188 & 0.157 & 0.148 & -0.045 & 0.670 \\
VAS+SAS (center, $\rho{=}0.20$) & 0.175 & 0.186 & 0.161 & 0.146 & -0.037 & 0.690 \\
VAS+SAS (center, $\rho{=}0.40$) & 0.198 & 0.227 & 0.175 & 0.133 & -0.042 & 0.705 \\
\bottomrule
\end{tabular}
\end{adjustbox}
\caption{\textbf{SAS sensitivity to projection and sparsification.} ``Dense'' keeps all projected coordinates; ``raw prompt residual''$^{\dagger}$ applies neither centering nor sparsification. The variants exhibit different trade-offs between whole-frame alignment and temporal structure, showing that projection and sparsification shape the residual rather than uniformly improving every metric.}
\label{tab:sas_ablation}
\end{table}

\begin{figure*}[t]
\centering
\includegraphics[width=0.9\textwidth]{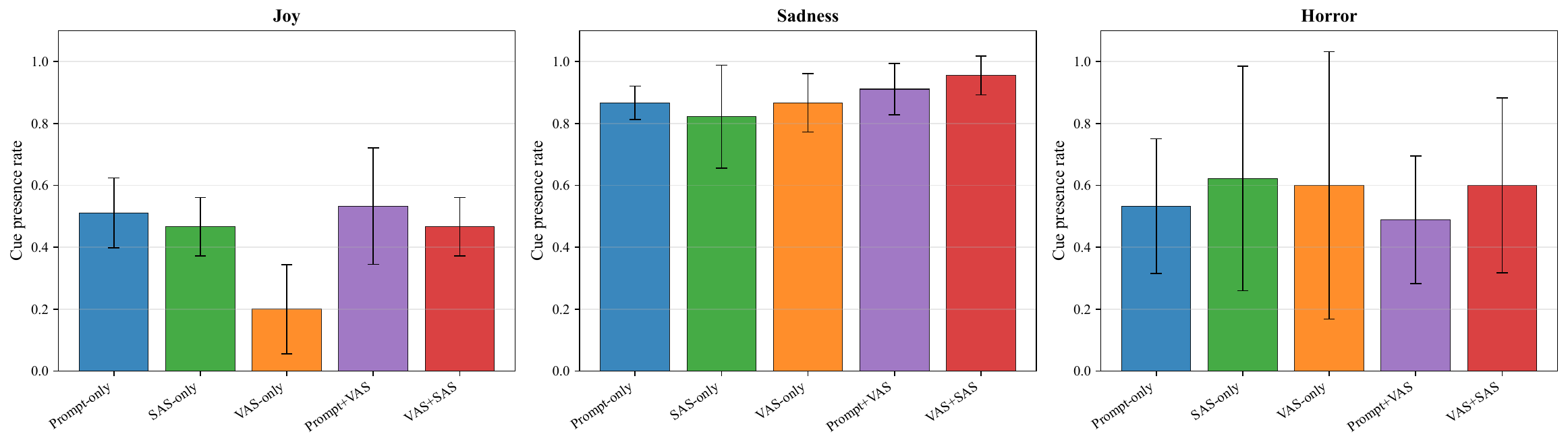}
\caption{\textbf{Semantic cue evaluation for SAS.} Grounding DINO measures the fraction of frames containing requested affect-bearing concepts. Cue+VAS and Cue+VAS+SAS use identical cue prompts and generation settings; their observed mean cue-presence rates are $\SASMatchedCueControl\%$ and $\SASMatchedCueFull\%$, respectively. The additional conditions illustrate the contributions of cue wording and residual routing.}
\label{fig:sas_cue_results}
\end{figure*}

\section{Focused Human Perceptual Validation}
\label{sec:human_study}

\paragraph{Study objective.}
The automatic metrics evaluate complementary aspects of
atmosphere, semantic-cue realization, and temporal affect
progression. We additionally conduct a focused human perceptual
study to test whether these operator-level improvements are
visible to viewers under matched pairwise comparisons. The study
is designed as perceptual validation of the three control
operators rather than as a comprehensive investigation of
cultural, demographic, or application-specific differences in
affective interpretation.

\paragraph{Participants and procedure.}
We recruited 60 participants and retained 54 after applying the
prespecified completion and attention-check criteria. Participant
demographics were a mean age of 27.4 years ($SD=6.8$, range
19--46); 28 participants identified as women, 25 as men, and 1 as
non-binary. Each participant completed 36 pairwise trials, yielding
1,944 valid judgments in total.

For every trial, two matched videos are displayed side by side
without method names. Left--right order and trial order are
randomized. The paired videos share the same scene, target emotion
or transition, prompt, initialization, seed, and sampling settings;
they differ only in the operator isolated by the corresponding
comparison. Participants are shown the target emotion labels and
task-specific instructions before making their judgments.
Table~\ref{tab:human_protocol} summarizes the three evaluation
tracks.

% Camera-ready: restore the exact approval/exemption status, institution, and protocol identifier.
\paragraph{Ethics and informed consent.}
The study protocol was reviewed and cleared under the authors' institutional
ethics process before data collection. The institution name and protocol
identifier are withheld solely to preserve double-blind anonymity and will be
disclosed in the camera-ready version. All participants provided informed
consent before beginning the study. Participation was voluntary, and
participants could withdraw at any time without penalty.

\begin{table*}[t]
\centering
\small
\setlength{\tabcolsep}{4.5pt}
\begin{adjustbox}{max width=\textwidth}
\begin{tabular}{llll}
\toprule
Component & Matched comparison & Primary perceptual question & Secondary perceptual question \\
\midrule
VAS
& Prompt-only vs. Prompt+VAS
& Which video better conveys the target atmosphere/emotion?
& Which video better preserves scene identity and overall visual quality? \\
SAS
& Cue+VAS vs. Cue+VAS+SAS
& Which video better realizes the requested affect-bearing semantic cues?
& Which video conveys the target affect more strongly and naturally? \\
TAS
& EFM (5-pass LERP) vs. TAS
& Which video shows a clearer progression from the start emotion to the end emotion?
& Which video presents a smoother and more temporally coherent transition? \\
\bottomrule
\end{tabular}
\end{adjustbox}
\caption{\textbf{Human perceptual-study protocol.}
Each track uses matched video pairs and task-specific questions
aligned with the intended control target. Method identities are
hidden from participants, and presentation order is randomized.}
\label{tab:human_protocol}
\end{table*}

\paragraph{Statistical analysis.}
For each task and question, we report the preference rate for the
EmoWorld configuration together with a 95\% confidence interval
obtained by resampling participants. The primary hypothesis test
compares each participant's mean preference with the 50\% chance
level using a two-sided participant-level permutation test. This
participant-clustered analysis avoids treating repeated judgments
from the same viewer as independent observations. The three
operator-level primary hypotheses (VAS, SAS, and TAS) form one
family, and we control the family-wise error rate across these three
permutation tests using the Holm--Bonferroni procedure. Reported
primary $p$-values are Holm-adjusted; secondary questions and their
confidence intervals are descriptive and are not included in this
multiplicity correction. All exclusion criteria, aggregation rules,
and analysis settings are fixed before examining the final study
outcomes.

\begin{table*}[t]
\centering
\small
\setlength{\tabcolsep}{5pt}
\begin{adjustbox}{max width=\textwidth}
\begin{tabular}{lllcc}
\toprule
Component
& Evaluated EmoWorld condition
& Primary preference [95\% CI]
& Secondary preference [95\% CI]
& Holm-adjusted primary $p$-value \\
\midrule
VAS
& Prompt+VAS
& 66.0\% [59.8\%, 72.1\%]
& 55.5\% [49.0\%, 61.9\%]
& $<0.003$ \\

SAS
& Cue+VAS+SAS
& 63.2\% [56.7\%, 69.5\%]
& 59.1\% [52.6\%, 65.3\%]
& $0.003$ \\

TAS
& TAS
& 71.4\% [65.6\%, 76.8\%]
& 67.0\% [60.9\%, 72.6\%]
& $<0.003$ \\
\bottomrule
\end{tabular}
\end{adjustbox}
\caption{\textbf{Human perceptual preferences.}
Primary and secondary questions follow
Table~\ref{tab:human_protocol}. Preference rates correspond to
the EmoWorld condition named in the second column. Confidence
intervals are computed by participant-level bootstrap, and primary
$p$-values are Holm--Bonferroni adjusted across the three operator-level
tests against the 50\% chance level. }
\label{tab:human_results}
\end{table*}

\paragraph{Results.}
For VAS, the observed preference rate for Prompt+VAS is 66.0\%
for target-atmosphere alignment (95\% CI [59.8\%, 72.1\%]) and
55.5\% for scene preservation and overall visual quality (95\% CI
[49.0\%, 61.9\%]; Holm-adjusted primary $p$-value $<0.003$). For SAS, the
observed preference rate for Cue+VAS+SAS is 63.2\% for
affect-bearing cue realization (95\% CI [56.7\%, 69.5\%]) and
59.1\% for overall target-affect communication (95\% CI
[52.6\%, 65.3\%]; Holm-adjusted primary $p$-value $=0.003$). For TAS, the
observed preference rate for TAS is 71.4\% for transition clarity
(95\% CI [65.6\%, 76.8\%]) and 67.0\% for transition smoothness
and temporal coherence (95\% CI [60.9\%, 72.6\%]; Holm-adjusted primary $p$-value $<0.003$).

These results provide a focused perceptual assessment of the
operator-specific effects quantified by the automatic evaluation.
Broader human-centered evaluation across annotator populations,
cultural contexts, affective interpretations, and application
settings remains an important direction for future work.

\section{Paired Affective Preparation Details}

\paragraph{Steering-vector observations.}
The VAS steering vectors are computed offline from paired neutral and emotion-conditioned panorama variants. For each selected layer and emotion, we average the corresponding feature difference and normalize the resulting steering vector before inference. No generator parameter, adapter, or auxiliary prediction network is optimized. The scene count, edit instructions, timestep and frame sampling, extraction hooks, and steering-vector version are recorded with the accompanying vectors.

\paragraph{Difference-aware cue mining.}
For each neutral and edited pair, frozen Qwen2.5-VL receives both images, the target emotion label, and a fixed instruction to report only affect-relevant visual changes while excluding unchanged content, geometry, and viewpoint. We parse the response into atomic atmosphere and semantic descriptors, normalize synonymous phrases, remove duplicates, and group the resulting descriptors by emotion. At inference, scene-compatible descriptors are retrieved from $\mathcal C_e$ and inserted into a fixed composition template together with the base scene prompt. The cue library, retrieval results, and final prompt pairs are stored in the accompanying manifest and frozen before matched comparisons.

\section{Emotion Taxonomy}
\label{sec:full_emotion_list}

We use 27 emotions based on the psychological taxonomy of Cowen and Keltner~\cite{cowen2017self}: Admiration, Adoration, Aesthetic Appreciation, Amusement, Anger, Anxiety, Awe, Awkwardness, Boredom, Calmness, Confusion, Craving, Disgust, Empathetic Pain, Entrancement, Excitement, Fear, Horror, Interest, Joy, Nostalgia, Relief, Romance, Sadness, Satisfaction, Sexual Desire, Surprise.

\section{Affective Cue Library and Prompt Composition}

Table~\ref{tab:semantic_prompts} gives six garden-scene examples obtained by composing the same base scene description with emotion-specific cues retrieved from the offline library. Cue+VAS and Cue+VAS+SAS reuse the exact same composed prompt in the matched comparison, so their difference is attributable to residual routing rather than prompt wording.

\begin{table*}[t]
\centering\small
\setlength{\tabcolsep}{4pt}
\renewcommand{\arraystretch}{0.95}
\begin{tabular}{lp{14cm}}
\toprule
Emotion & Emotion-augmented prompt \\
\midrule
Calmness & A peaceful garden with soft morning light filtering through leaves, gentle breeze, a clear reflective pond with lily pads, stone pathway winding through greenery \\
Horror & A dark abandoned garden with twisted thorny branches, a murky stagnant pond reflecting pale moonlight, withered blackened flowers, fog creeping between dead trees \\
Joy & A vibrant sunny garden with colorful blooming flowers, bright golden sunlight, butterflies in warm air, a sparkling fountain with rainbow mist \\
Sadness & A wilted garden in cold gray rain, drooping flowers heavy with raindrops, a dried-up pond with cracked mud, bare branches reaching like empty hands \\
Fear & A garden shrouded in darkness with looming shadows, rustling sounds from unseen corners, a dark pond with something lurking beneath, thorny vines creeping closer \\
Excitement & A dynamic garden in dramatic sunset light, flowers swaying wildly in strong wind, splashing fountain with vibrant energy, vivid clouds and colors \\
\bottomrule
\end{tabular}
\caption{\textbf{Examples of cue-library-composed SAS prompts for one scene.} The cue library spans the complete 27-emotion taxonomy.}
\label{tab:semantic_prompts}
\end{table*}

\section{Taxonomy-Wide Affective Control}
\label{sec:per_emotion}

This section reports the complete 27-emotion results that complement the standardized cross-system comparison in Table~\ref{tab:single_emotion}. Category-level scores characterize the response distribution across the taxonomy, while method-level conclusions are drawn from macro-aggregated results and matched protocols.

\begin{table*}[t]
\centering\small
\setlength{\tabcolsep}{3pt}
\begin{adjustbox}{max width=\textwidth}
\begin{tabular}{lcccccccc}
\toprule
& \multicolumn{4}{c}{T2V} & \multicolumn{4}{c}{I2V} \\
\cmidrule(lr){2-5} \cmidrule(lr){6-9}
Emotion & No-St. & Prompt & +VAS & $\Delta$ & No-St. & Prompt & +VAS & $\Delta$ \\
 & CLIP-Emo & CLIP-Emo & CLIP-Emo & & CLIP-Emo & CLIP-Emo & CLIP-Emo & \\
\midrule
Admiration & 0.192 & 0.210 & 0.212 & +0.001 & 0.172 & 0.173 & 0.192 & +0.019 \\
Adoration & 0.141 & 0.161 & 0.168 & +0.007 & 0.158 & 0.161 & 0.178 & +0.018 \\
Aesthetic & 0.197 & 0.199 & 0.199 & +0.000 & 0.182 & 0.174 & 0.182 & +0.008 \\
Amusement & 0.169 & 0.138 & 0.200 & +0.061 & 0.203 & 0.220 & 0.139 & -0.081 \\
Anger & 0.141 & 0.188 & 0.183 & -0.005 & 0.152 & 0.157 & 0.196 & +0.039 \\
Anxiety & 0.143 & 0.143 & 0.187 & +0.043 & 0.180 & 0.164 & 0.188 & +0.023 \\
Awe & 0.184 & 0.206 & 0.202 & -0.004 & 0.164 & 0.178 & 0.174 & -0.004 \\
Awkwardness & 0.131 & 0.166 & 0.163 & -0.003 & 0.160 & 0.145 & 0.158 & +0.013 \\
Boredom & 0.188 & 0.193 & 0.198 & +0.005 & 0.214 & 0.201 & 0.189 & -0.012 \\
Calmness & 0.157 & 0.171 & 0.181 & +0.010 & 0.161 & 0.125 & 0.152 & +0.027 \\
Confusion & 0.169 & 0.172 & 0.208 & +0.037 & 0.205 & 0.196 & 0.203 & +0.007 \\
Craving & 0.140 & 0.147 & 0.164 & +0.017 & 0.177 & 0.147 & 0.170 & +0.023 \\
Disgust & 0.182 & 0.198 & 0.194 & -0.004 & 0.211 & 0.221 & 0.195 & -0.026 \\
Emp. Pain & 0.137 & 0.138 & 0.137 & -0.001 & 0.179 & 0.163 & 0.179 & +0.016 \\
Entrancement & 0.207 & 0.203 & 0.212 & +0.009 & 0.184 & 0.187 & 0.168 & -0.018 \\
Excitement & 0.150 & 0.103 & 0.163 & +0.060 & 0.180 & 0.176 & 0.191 & +0.015 \\
Fear & 0.149 & 0.169 & 0.226 & +0.057 & 0.175 & 0.169 & 0.180 & +0.011 \\
Horror & 0.143 & 0.147 & 0.218 & +0.071 & 0.150 & 0.176 & 0.177 & +0.001 \\
Interest & 0.125 & 0.116 & 0.149 & +0.033 & 0.165 & 0.172 & 0.149 & -0.022 \\
Joy & 0.141 & 0.059 & 0.163 & +0.105 & 0.167 & 0.147 & 0.184 & +0.037 \\
Nostalgia & 0.139 & 0.104 & 0.148 & +0.045 & 0.118 & 0.102 & 0.125 & +0.023 \\
Relief & 0.151 & 0.130 & 0.141 & +0.011 & 0.161 & 0.153 & 0.161 & +0.008 \\
Romance & 0.133 & 0.108 & 0.166 & +0.058 & 0.113 & 0.119 & 0.138 & +0.019 \\
Sadness & 0.162 & 0.157 & 0.234 & +0.077 & 0.189 & 0.189 & 0.184 & -0.004 \\
Satisfaction & 0.170 & 0.148 & 0.191 & +0.043 & 0.166 & 0.188 & 0.184 & -0.004 \\
Sex. Desire & 0.154 & 0.137 & 0.136 & -0.001 & 0.155 & 0.170 & 0.155 & -0.015 \\
Surprise & 0.202 & 0.174 & 0.207 & +0.033 & 0.206 & 0.194 & 0.220 & +0.026 \\
\midrule
\textbf{Macro Avg} & 0.159 & 0.155 & 0.183 & +0.028 & 0.172 & 0.169 & 0.174 & +0.005 \\
\bottomrule
\end{tabular}
\end{adjustbox}
\caption{\textbf{Per-emotion CLIP-Emo scores on the full 27-emotion benchmark.} $\Delta$ is the improvement of Prompt+VAS over Prompt-only. All videos use 49 frames at $832\times480$.}
\label{tab:per_emotion_27}
\end{table*}

\begin{figure*}[t]
\centering
\includegraphics[width=0.95\textwidth]{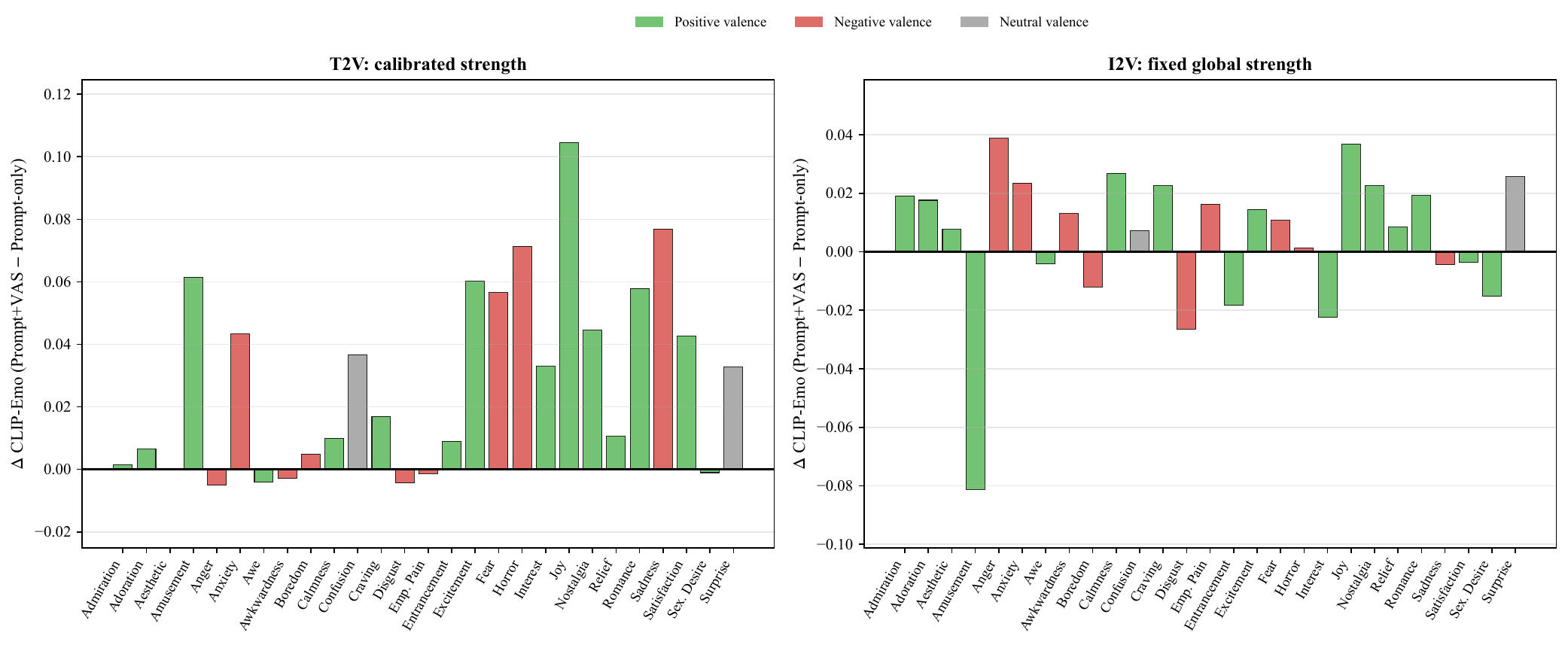}
\caption{\textbf{Category-wise VAS gains across the complete 27-emotion taxonomy.} Bars show $\Delta$CLIP-Emo relative to Prompt-only under the calibrated T2V and fixed-strength I2V protocols. The distributions complement the macro-aggregated evaluation by revealing category-dependent response patterns; bar colors indicate valence grouping.}
\label{fig:emotion_27_gain}
\end{figure*}
\begin{figure*}[t]
\centering
\includegraphics[width=0.95\textwidth]{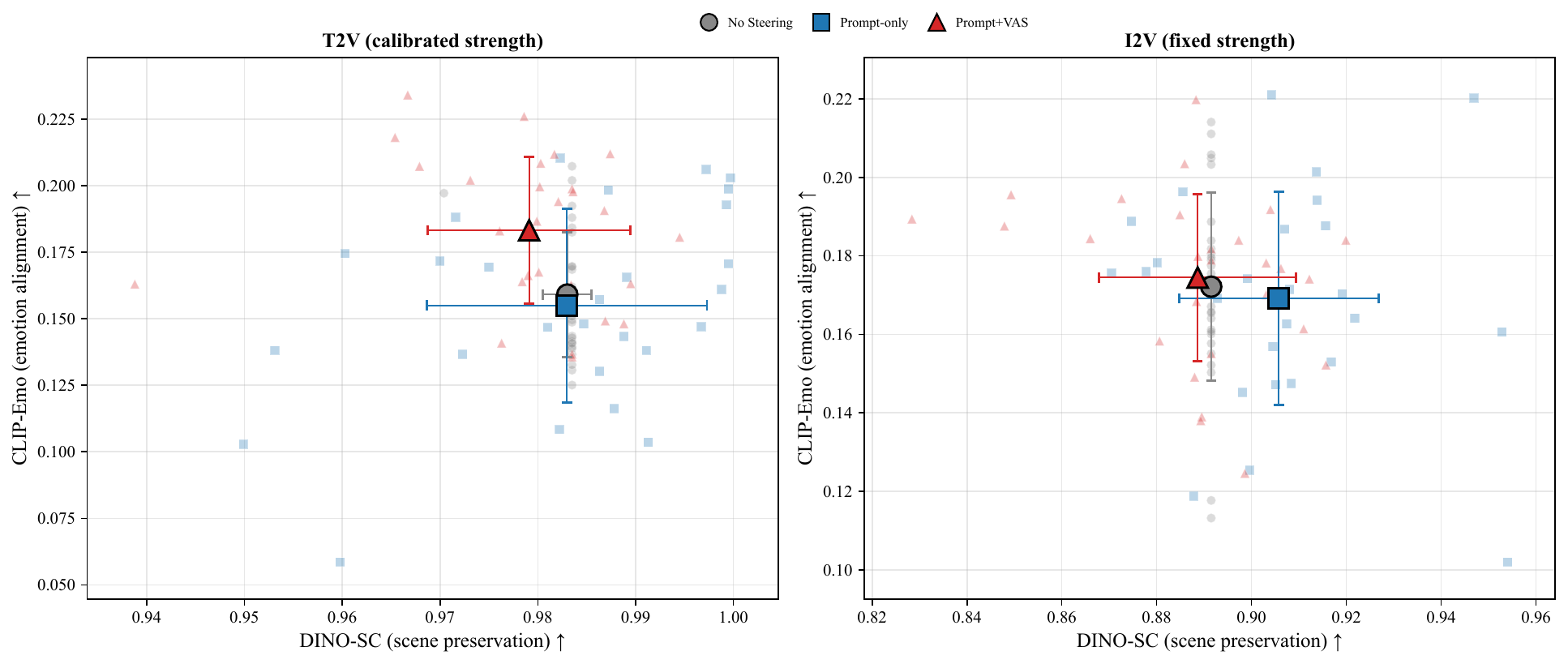}
\caption{\textbf{Alignment--preservation analysis on the complete 27-emotion benchmark.} Each point represents one evaluated emotion--scene instance, and large markers show macro averages with standard-deviation error bars. T2V uses per-emotion calibrated steering strength (DINO-SC $0.983{\to}0.979$, $\Delta{-}0.004$); I2V uses fixed strength (DINO-SC $\geq 0.88$ across methods). The trade-off curve shows that calibrated T2V and fixed I2V both retain structural fidelity while gaining emotion alignment.}
\label{fig:alignment_preservation}
\end{figure*}

Figures~\ref{fig:emotion_27_gain} and~\ref{fig:alignment_preservation} show the category-wise gains and the corresponding alignment--preservation trade-off. The aggregate affect gain is obtained with only a small change in structural consistency under the corresponding protocols.

\subsection{Generalization Across Settings and Backbones}

To isolate method-level generalization from category-specific variation, Table~\ref{tab:sup_cross_setting_generalization} macro-averages the four representative emotions with equal category weight. We compare the same steering principle across T2V and I2V generation on Wan2.2-5B and further instantiate it
on CogVideoX-5B using backbone-specific steering vectors and architecture-compatible feature hooks. Category-level variation is analyzed more comprehensively in the complete 27-emotion benchmark, rather than being treated as a separate factor in this matched cross-setting comparison.

\begin{table*}[t]
\centering
\small
\setlength{\tabcolsep}{5.5pt}
\renewcommand{\arraystretch}{1.08}
\begin{tabular}{lllccc}
\toprule
Mode & Backbone & Method
& CLIP-Emo $\uparrow$
& EI $\uparrow$
& EQI $\uparrow$ \\
\midrule

\multirow{4}{*}{T2V}
& \multirow{4}{*}{Wan2.2-5B}
& No steering
& 0.154 & -0.068 & 0.024 \\
&
& Prompt-only
& 0.168 & -0.049 & 0.029 \\
&
& RAVE
& 0.166 & -0.055 & 0.032 \\
&
& Prompt+VAS (Ours)
& \textbf{0.200} & \textbf{-0.026} & \textbf{0.035} \\

\midrule

\multirow{3}{*}{I2V}
& \multirow{3}{*}{Wan2.2-5B}
& No steering
& 0.159 & -0.063 & 0.025 \\
&
& Prompt-only
& 0.162 & -0.060 & 0.025 \\
&
& Prompt+VAS (Ours)
& \textbf{0.171} & \textbf{-0.053} & \textbf{0.027} \\

\midrule

\multirow{3}{*}{I2V}
& \multirow{3}{*}{CogVideoX-5B}
& No steering
& 0.146 & -0.056 & 0.025 \\
&
& Prompt-only
& 0.148 & -0.054 & 0.026 \\
&
& Model-specific VAS (Ours)
& \textbf{0.153} & \textbf{-0.051} & \textbf{0.028} \\

\bottomrule
\end{tabular}
\caption{\textbf{Method-level generalization across generation settings and Video-DiT backbones.} Values are macro-aggregated over the standardized shared-protocol subset, using the same aggregation as Table~\ref{tab:single_emotion}. Wan Prompt-only and Prompt+VAS use identical composed prompts; CogVideoX uses independently extracted steering vectors and architecture-compatible hooks.}
\label{tab:sup_cross_setting_generalization}
\end{table*}

\begin{figure*}[t]
    \centering
    \includegraphics[width=0.96\textwidth]{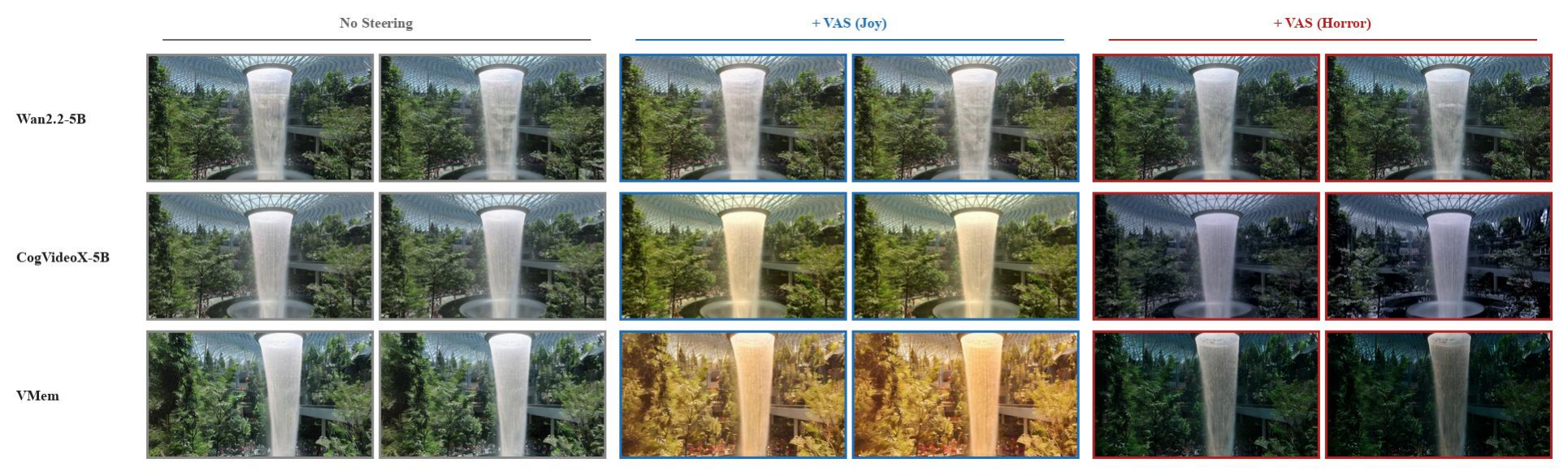}
    \caption{\textbf{Cross-backbone qualitative portability.} Backbone-specific VAS vectors instantiate the same extraction--injection principle on Wan2.2-5B, CogVideoX-5B, and VMem across matched scene and affect conditions. The columns show no steering and VAS steering toward joy or horror; SAS and TAS are inactive.}
    \label{fig:cross_model_qual}
\end{figure*}

Across the three settings, VAS improves aggregate emotion alignment and emotion margin over the corresponding Prompt-only condition. The largest gain occurs in Wan T2V, where internal steering has greater freedom to establish global atmosphere without a reference-image constraint. Positive aggregate gains persist in Wan I2V, and the CogVideoX results show that the extraction--injection principle remains portable when vectors and hook locations are instantiated for a different Video-DiT architecture. Figure~\ref{fig:cross_model_qual} provides the corresponding qualitative comparison across backbones.

% The previous overloaded raster gallery (fig2/supp_gallery.pdf) is deliberately
% not included. Add the redesigned single-page vector/mixed-vector gallery below
% only after it passes the visual QA described in the server-agent README.

\section{Additional Qualitative Gallery}
\label{sec:supp_gallery}
\begin{figure*}[t]
    \centering
    \includegraphics[width=\textwidth]{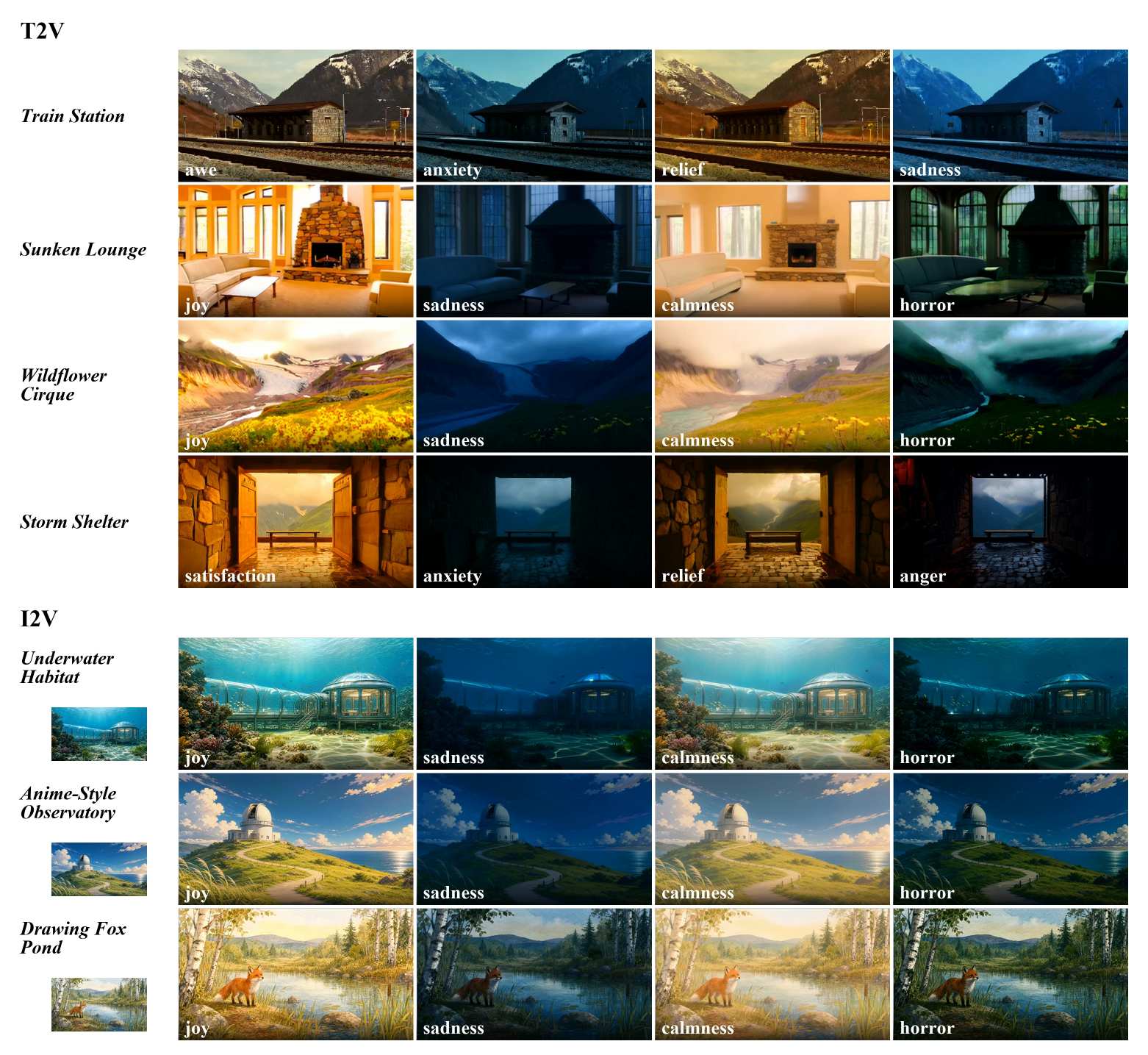}
    \caption{\textbf{Curated qualitative gallery.} Representative T2V and I2V atmosphere-control examples are organized by scene and labeled with target emotion.}
    \label{fig:supp_gallery_1}
\end{figure*}

\begin{figure*}[t]
    \centering
    \includegraphics[width=\textwidth]{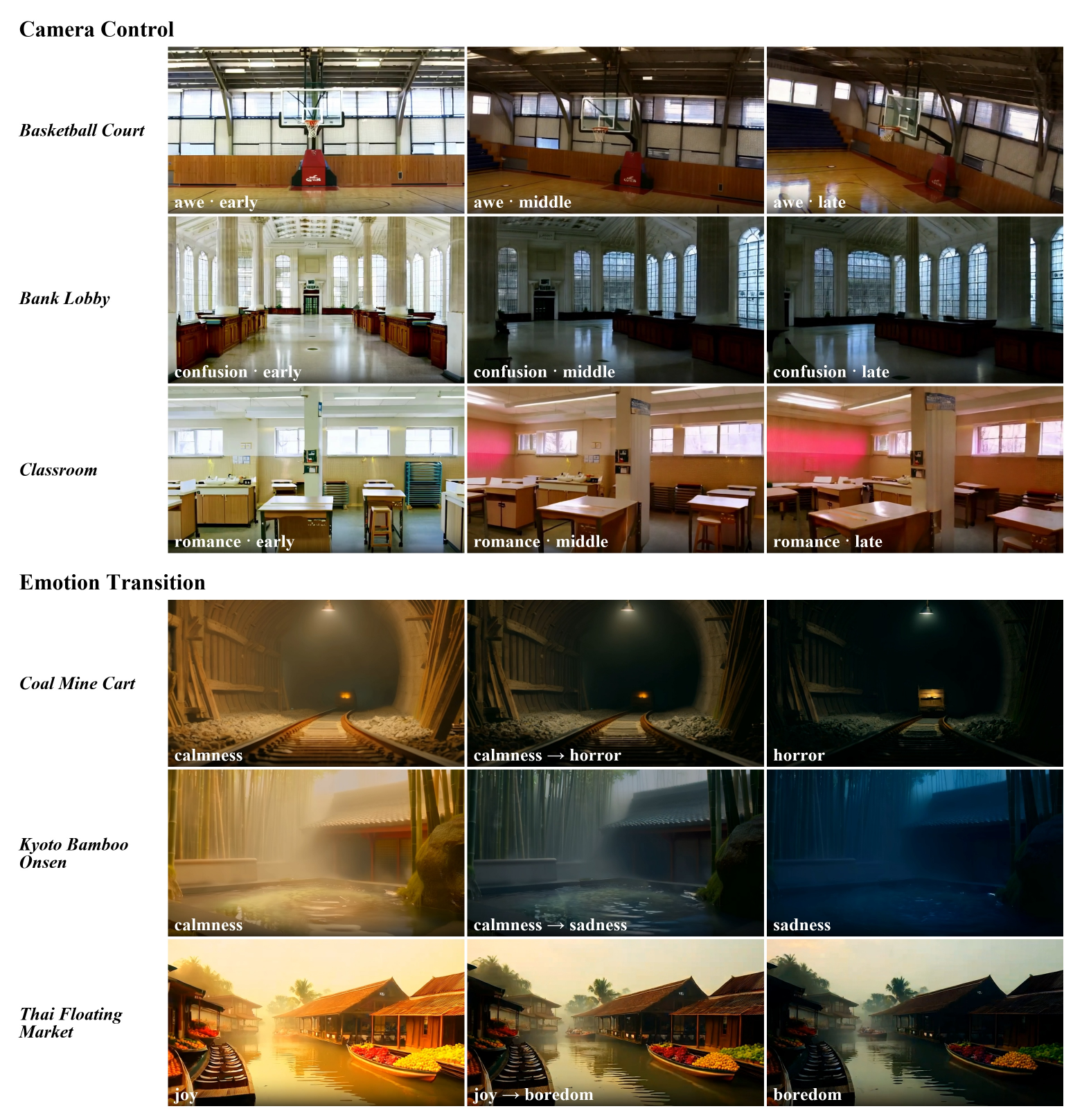}
    \caption{\textbf{Curated qualitative gallery.} Temporal-transition and camera-conditioned examples show ordered RGB frames along the generated affect and viewpoint trajectories.}
    \label{fig:supp_gallery_2}
\end{figure*}

Figures~\ref{fig:supp_gallery_1} and~\ref{fig:supp_gallery_2} collect additional qualitative examples across the supported control settings. The gallery
includes T2V and I2V atmosphere control over diverse scenes and
emotion categories, temporal transitions between endpoint
emotions, and camera-conditioned generation that combines
prescribed viewpoint trajectories with scheduled affect steering.
These examples provide a broader qualitative view of EmoWorld's
control capabilities and their composition across different
generation settings.

\clearpage

\end{document}